\documentclass[conference]{IEEEtran}
\usepackage{amsmath,amssymb,amsfonts, amsthm}
\usepackage{algpseudocode, algorithm}
\usepackage{graphicx}
\usepackage{caption}
\usepackage{subfigure}
\usepackage{wrapfig}
\usepackage{xcolor}
\usepackage[hyphens]{url}
\usepackage{hyperref}
\usepackage[hyphenbreaks]{breakurl}
\usepackage{booktabs}       
\usepackage{nicefrac}       
\usepackage{microtype}      
\usepackage{textcomp}
\usepackage{bookmark}
\usepackage{mathtools}

\usepackage{makecell}
\usepackage{xspace}

\usepackage{epsfig,endnotes}
\usepackage{multirow}

\usepackage{tablefootnote}
\usepackage{hyperref}       

\theoremstyle{plain}
\newtheorem{theorem}{Theorem}[section]

\theoremstyle{definition}
\newtheorem{definition}[theorem]{Definition}

\theoremstyle{remark}

\newcommand{\model}{{\mathcal{M}}}

\newcommand{\trans}{{\mathcal{AT}}}

\newcommand{\modelori}{{\mathcal{M}_{ori}}}
\newcommand{\featurespaceori}{{\mathcal{F}_{ori}}}

\newcommand{\areafunc}{{\mathcal{S}}}

\newcommand{\ylim}{{y_{\text{lim}}}}
\newcommand{\xcenter}{{c}}
\newcommand{\totalversions}{{N}}

\newcommand{\trainori}{{D_{\text{train}}}}
\newcommand{\testori}{{D_{\text{test}}}}
\newcommand{\trainhidden}{{D_{\text{hidden}}}}

\newcommand{\hiddendata}{\mathcal{X}_h}
\newcommand{\hiddenfeature}{h}

\newcommand{\attackr}{{\mathcal{AR}}}

\newcommand{\revision}[1]{{\color{black} #1}}
\newcommand{\arevision}[1]{{\color{black} #1}}

\newcommand{\para}[1]{{\vspace{4pt} \bf \noindent #1 \hspace{6pt}}}

\newcommand{\cifar}{{\tt CIFAR10}}
\newcommand{\ytface}{{\tt YouTube Face}}
\newcommand{\skincancer}{{\tt Skin Cancer}}

\newcommand{\eg}{{\em e.g.,\ }}
\newcommand{\ie}{{\em i.e.,\ }}

\newenvironment{packed_itemize}{
\begin{list}{\labelitemi}{\leftmargin=0.5em}
  \setlength{\itemsep}{2pt}
  \setlength{\parskip}{0pt}
  \setlength{\parsep}{0pt}
  \setlength{\headsep}{0pt}
  \setlength{\topskip}{0pt}
  \setlength{\topmargin}{0pt}
  \setlength{\topsep}{0pt}
  \setlength{\partopsep}{0pt}
}{\end{list}}

\DeclareMathOperator*{\argmin}{arg\,min}

\begin{document}

\title{Towards Scalable and Robust Model Versioning}
\author{Wenxin Ding, Arjun Nitin Bhagoji, Ben Y. Zhao, Haitao Zheng\\
{\em Department of Computer Science, University of Chicago}\\
{\em \{wenxind, abhagoji, ravenben, htzheng\}@cs.uchicago.edu}}

\maketitle

\begin{abstract}

As the deployment of deep learning models continues to expand across industries, the threat of malicious incursions aimed at gaining access to these deployed models is on the rise. Should an attacker gain access to a deployed model, whether through server breaches, insider attacks, or model inversion techniques, they can then construct white-box adversarial attacks to manipulate the model's classification outcomes,  thereby posing significant risks to organizations that rely on these models for critical tasks.  Model owners need mechanisms to protect themselves against such losses without the necessity of acquiring fresh training data - a process that typically demands substantial investments in time and capital. 

In this paper, we explore the feasibility of generating multiple versions of a model that possess different attack properties, without acquiring new training data or changing model architecture. The model owner can deploy one version at a time and replace a leaked version immediately with a new version. The newly deployed model version can resist adversarial attacks generated leveraging white-box access to one or all previously leaked versions. We show theoretically that this can be accomplished by incorporating parameterized {\em hidden distributions} into the model training data, forcing the model to learn task-irrelevant features uniquely defined by the chosen data.  Additionally, optimal choices of hidden distributions can produce a sequence of model versions capable of resisting compound transferability attacks over time. Leveraging our analytical insights, we design and implement a practical model versioning method for DNN classifiers, which leads to significant robustness improvements over existing methods. We believe our work presents a promising direction for safeguarding DNN services beyond their initial deployment.

\end{abstract}

\section{Introduction} 
\label{sec:intro}

As deep learning models become increasingly prevalent across various
industries, the risk of malicious attacks attempting to breach access
to these deployed models is growing.  When an attacker obtains
access to a deployed model, via many methods including server breaches,
insider attacks, or model inversion attacks, they can craft white-box
adversarial attacks to manipulate the
model's classification results. As the classification results lose
their reliability, the deployed ML service becomes obsolete and needs
to be replaced or removed. \revision{For example, in the medical
  field, image classification faces real-world attacks that aid 
  insurance fraud.  Some physicians engage in
  overprescribing medications and misdiagnosing conditions, prompting
  insurance companies to employ image classification for diagnosis
  verification.  Adversarial attacks against these models are
  well-documented and raise significant concerns~\cite{finlayson2019adversarial,
    MA2021}. Another example is classifiers for image content
  moderation on discussion boards and online platforms, which are
  frequently attacked by bad actors seeking to bypass content
  moderation with prohibited content~\cite{RAHMAN2023103583,bhagoji2018practical}.
}

Unfortunately, replacing a breached model is a challenging task. This
is because,  building powerful deep learning models often involves acquiring and curating
high quality training datasets, a process that incurs significant investment
in time and capital, and can be difficult or impractical to repeat. For
example, a model identifying \revision{diseases like rare skin lesions} may take years to collect
the training data, involving curation by specialists and
de-identification to comply with privacy
regulations~\cite{tschandl2018ham10000}.  \revision{Similarly,
  training data for content moderation models is often extremely
  sensitive and difficult to procure, e.g. images of extreme violence
  or abuse of minors. Obtaining and labeling these challenging images require manual inspection and careful supervision.}

\revision{These observations motivate us to} investigate how model owners can protect themselves
against model losses {\em without} the
necessity of acquiring new training data. 
Ideally, they would like to train multiple versions of the target model from the same
training set,  deploy one version at a time and replace a leaked
version with a new one.  At the time of
its deployment, each new version $i$
of the model should resist adversarial attacks as if it were the {\em
 sole} model version that
had been deployed. That is, when attempting to attack model $i$,  an attacker would gain
minimal or no advantage by obtaining white-box access to any or  all
previous model versions $\{j \; | \; j<i\}$.  The longer the model sequence 
supports this {\em sequential} (or {\em compound}) robustness, the stronger the
protection against repeated model leakages.   We refer to this problem as {\em scalable and robust
  model versioning}.

Unfortunately, solving this problem is difficult because of two significant
challenges.  First is the well known phenomenon of {\em attack
  transferability}: adversarial attacks generated on one model often succeed
on similar models trained on the same task, even when they use different
architectures~\cite{demontis2019adversarial,liu2016delving}.  To date, few if
any techniques provide the ability to generate multiple ($>3$) models on the
same task with low attack transferability between them, while maintaining
normal classification accuracy.  Second, our problem must address a novel
but natural evolution of the white-box attack, {\em a sequential,
  compound transferability attack}. As each leaked model is retired, an attacker with
no white-box access to the new version $i$, can still utilize their white-box
access to prior versions ($1$ to $i-1$) to orchestrate a strong attack
against version $i$. This new requirement and the need to sequentially deploy/replace
models make our problem distinct from existing works (\eg constructing
robust ensembles~\cite{yang2020dverge,yang2021trs,pmlr-v162-dbouk22a}),
creating a new, open challenge.

\revision{To tackle these challenges, our work introduces a principled 
  approach for generating a sequence of robust model versions from a
  {\em single} training dataset using a {\em single} model
  architecture.   Our method automatically curates and
  incorporates {\em parameterized hidden data} into the model training data,
  forcing the model to learn task-irrelevant features that are distinctly defined by the selected hidden data.
More specifically, given an original task,  we curate synthetic data that
are drawn from some hidden distributions irrelevant to the
task,   and augment the task's original training data with the new
hidden data.  Thus, the training data of a class now includes both its original training data and
 new data drawn from a chosen hidden distribution. Using the combined
 data, we train a model
 version from scratch. And by varying the choices of hidden
distributions, we can produce
different model versions. }

\revision{Our hypothesis is that the above process, if well-designed,
  can produce a diverse set of model versions. These model versions
  would not only effectively accomplish the primary task but also
  exhibit distinct attack properties, because they are trained on
  varying hidden distributions that naturally introduce diverse
  non-robust features into each model.  Leveraging this variability,
  the model trainer can carefully select, organize and continuously deploy a {\em sequence} of model
  versions to withstand compound transferability
  attacks over time.  
} 

\revision{We validate this hypothesis by first performing theoretical
  analysis in a simplified setting.  Our formal proof demonstrates that optimizing 
 the selection of hidden distributions significantly reduces the
 transferability of compound
 attacks against subsequent model versions.   Additionally,  effective hidden
  distributions are characterized by a single point in the feature space,
  facilitating the parameterization process for efficient
  optimization. Together,  these analytical findings demonstrate the
  importance and viability of optimizing the choices of hidden distributions when constructing
  the model sequence. This stands in contrast to an earlier work~\cite{shan2022post}  that
  employed randomly selected hidden data without any optimization. 
  
Building on our analytical results, we 
  develop a practical,  greedy search-based algorithm for constructing hidden distribution-based
  model versions for deep neural network (DNN) classifiers.  We
  implement and evaluate our method using three image
  classification tasks that encompass different image categories
  (objects, medical images, and faces) and varying number of classes
  (7  to 1283). Our design significantly outperforms alternative methods for model
  versioning, including \cite{shan2022post}  and those designed to
  produce ``orthogonal'' models.

\para{Summary of Contributions.} To the best of our knowledge, our
work is the first to provide a principled investigation of scalable and robust model
versioning -- a crucial
task for safeguarding DNN services {\em beyond} their initial
deployment.  Our work makes three key contributions: 
\begin{packed_itemize}
\item We formally define the process of hidden
distribution-based  
training as a solution for model versioning
(\S\ref{sec:problem} and \S\ref{sec:hiddentrain});
\item We analytically demonstrate the critical impact of
    hidden distributions on model versioning and develop a practical
    algorithm for systematically selecting hidden distributions to
    construct robust model versions (\S\ref{sec:theory} and
    \S\ref{sec:system}).  
  \item We evaluate our design by building a sequence of model
    versions for three image classification tasks.  These models
    achieve significantly higher robustness against attacks compared to existing methods (\S\ref{sec:exp}). 
\end{packed_itemize}
Finally, we discuss the limitations of our work and potential
future directions.  We hope that our work can inspire further
research efforts in this critical yet underexplored area. 

}

\vspace{0.05in}
\section{Background and Related Work}
\label{sec:background}

To provide context, we briefly describe transferability of adversarial example attacks,  existing methods to restrict transferability between models, and those to produce model variants.

\label{subsec:advexp}
\para{Transferability of Adversarial Examples.} It is well-known that adversarial examples generated for one model can produce misclassification on other similar models~\cite{goodfellow2014explaining,szegedy2013intriguing}, a phenomenon known as attack \emph{transferability}.  It is widely used by attacks where attackers have no access to the internals of the target model. One can also increase attack transferability by modifying attack methodology~\cite{liu2016delving,wu2021improving}, gaining limited query access to the target model~\cite{suya2020hybrid} and leveraging learned model features~\cite{inkawhich2020transferable,springer2021little}. When exploring conditions that produce high attack transferability,  Demontis et al. showed that the alignment of input gradients between models and the reduction of variability of the loss surface are critical for high transferability~\cite{demontis2019adversarial}.  Others found that transferability between models correlates strongly with 
their semantic layer similarity~\cite{cianfarani2022understanding} or feature similarity~\cite{WIEDEMAN2022}.

\para{Reducing (Pairwise) Attack Transferability.}   One can apply
adversarial training (\eg\cite{madry2017towards}) to reduce a model's
vulnerability to attacks, which may help reduce attack transferability from
other models.  However, doing so faces high training cost and reduced task
accuracy.  Recent works (\eg DVERGE~\cite{yang2020dverge} and
TRS~\cite{yang2021trs}) proposed to train an ensemble of diverse models with low
pairwise attack transferability, to resist adversarial attacks against the
entire ensemble.  These ensemble methods, like adversarial training, require
iterative optimization to simultaneously train the full ensemble at once. The resulting 
optimization is computationally expensive, especially for ensembles with more
than 3 models.  For example, DVERGE~\cite{yang2020dverge} reported a high
pairwise attack transferability of 79\% when training an ensemble of size 5.

Our work considers a very different problem.  Rather than training a set of  models that operate as an ensemble to classify inputs and resist attacks (where attackers know either all or none of the models), we seek to produce a sequence of model versions and deploy them one by one in a sequential order, each triggered by a model leakage event.  Each model version $i$ operates by itself to classify inputs and faces a new type of black-box attacks constructed using all prior versions.  That is, while an attacker has no access to version $i$,  it can leverage white-box access to all prior versions ($1$ to $i-1$) to build a powerful ``group-based''  attack against version $i$. Previous studies have not considered this novel form of attack, and mitigating it cannot be achieved solely by reducing pairwise attack transferability.

\para{Model Versioning.}  
\label{subsec:modelver}
In practice, deployed models often evolve over time to adapt to changes in data or to incorporate new training methods~\cite{MLops,StableDiffusion,xu2021deep}.  However, none of these methods explicitly consider how to update a model after it is leaked to attackers.   The only known work on this topic is \cite{shan2022post}, which randomly selects hidden distributions to produce model variants.
\revision{However, \cite{shan2022post} does not consider reducing attack transferability, instead focusing on creating a separate input filter to identify potential adversarial examples at run-time.  Our work is inspired by \cite{shan2022post},  but differs in two key aspects. First, we study whether and how hidden distributions can be optimized to minimize attack transferability against subsequent model versions. We are the first to establish an analytical framework to formally illustrate the critical impact and viable path for optimizing the selection of hidden distributions. Notably, our findings stand in contrast to \cite{shan2022post}'s random selection decision.   Second, we develop a principled method for selecting hidden distributions and organizing a sequence of model versions for DNN classifiers.  Experiments in \S\ref{sec:exp} show that our method significantly outperforms \cite{shan2022post} both with and without the input filter. Therefore, our study makes a tangible step forward in tackling the challenging problem of model versioning.}

\para{Detecting Model/Server Breaches.}  One assumption made by our work is that the model owner/deployer can detect that the current model version has been breached, so that our work can focus on the problem of finding replacement models. This assumption is not unrealistic.  Many years of research in the computer security community has focused on intrusion detection and on detecting when a server has been breached (and when specific files are accessed), \eg intrusion detection systems, APT detection/analysis, and OS-level secure access logs. In addition, classification systems rarely stand alone, and often lead to downstream errors and negative outcomes. Any attacker that uses the breached model to trigger negative outcomes downstream can be detected, with the downstream errors used to trace back to the compromised model, \eg post-attack forensic systems~\cite{shan2022poison}.

\section{Problem and Threat Model} 
\label{sec:problem}
In this section, we describe the problem of model versioning and its threat model. We focus our discussion on the adversary's capabilities once they have breached the deployed model and the objectives of the model owner in responding to and recovering from such a breach.   We start from the simple scenario of single (or one-time) model breach and then move to the broader context of multiple (or repeated) model breaches.  With these considerations in mind, we also examine existing solutions and their limitations.

For the rest of the paper, we consider the standard multi-class classification model $\model: \mathcal{X} \subseteq \mathbb{R}^n \to \mathcal{L}$.
The classification task is the main task where 
training data on the main task is denoted as $\trainori$ and test data on the main task is $\testori$.

\subsection{One-time Breach}
\label{subsec:one_breach} 

We first consider the simple case where a deployed model $\model_1$ is breached by an attacker. Here our work assumes that the model owner/deployer has detected the breach.  To recover from such loss and continue the model service, the model owner replaces $\model_1$ with a new version $\model_2$.

\begin{figure}[t]
  \centering
 \includegraphics[width=0.2\linewidth, angle=270]{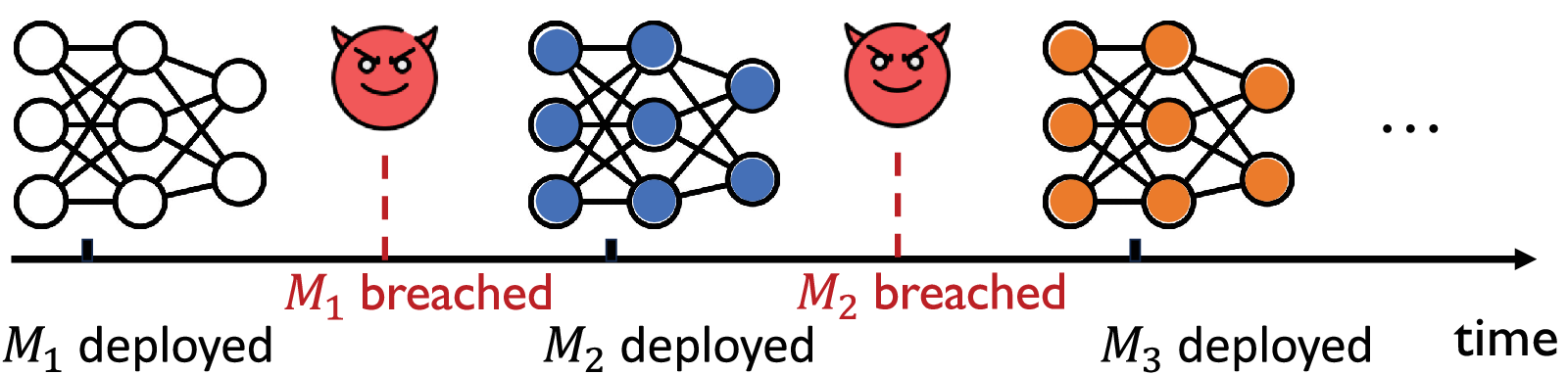}
 \caption{\em Model versioning: after the deployed model $\model_i$ is breached by the attacker, the model owner replaces it with a new version $\model_{i+1}$ to ensure uninterrupted service.}
  \label{fig:version}
   \vspace{-0.15in}
\end{figure}

\para{Adversary Capabilities.} In this case, the parameters of $\model_1$ are, at some point, leaked or
inferred by an attacker (see Figure~\ref{fig:version}). While several threats to data privacy and security
could arise from such a leak, we focus on the particularly pernicious one of
\emph{white-box adversarial examples}.
Here we assume the attacker is {capable} of
constructing targeted adversarial examples
\footnote{We focus on targeted adversarial examples for ease of exposition. All
results hold for untargeted adversarial examples as well.} with the knowledge of $\model_1$'s architecture
and weights. 
Specifically, for a desired class $l_t \in \mathcal{L}$, the attacker can generate $\tilde{x}=x+\delta_x$ such that 
\[\model_1(\tilde{x})=l_t \neq \model_1(x)\] with $d(x,\tilde{x})<\epsilon$, where $l_t$ is not the correct class of $x$ and $d(\cdot,\cdot)$ is a distance function (usually a $p$-norm). Finally, since the attacker has no knowledge of the replacement model $\model_2$, they will generate white-box
adversarial examples from $\model_1$ and leverage transferability to attack $\model_2$.

\para{Model Owner's Goal.}  To recover from the breach, the model owner deploys a new model $\model_2$ to replace $\model_1$, where 
\begin{packed_itemize}
  \item $\model_2$ maintains high performance on the main task assessed by the test data $\testori$;
  \item $\model_2$ is {robust} against white-box adversarial attacks computed based on $\model_1$.
\end{packed_itemize}

\para{Robustness Measure.}  Here the threat to the model owner is the leakage of $\model_1$, which the attacker can leverage to craft white-box adversarial examples.   Thus, we measure the robustness of the model replacement by the {\bf {\em directional attack transferability}} from $\model_1$ to $\model_2$: \[\trans (\model_1 \rightarrow \model_2).\]  This metric is directional, \ie the sequential order of the model deployment matters.

\subsection{Multiple Breaches}
\label{subsec:multiple_breaches}

In practice, model breaches can happen multiple times -- after replacing $\model_1$ with $\model_2$,  the deployed model $\model_2$ can be breached, at some point, forcing the model owner to replace it with $\model_3$, etc. (see Figure~\ref{fig:version}).  And more importantly, after breaching the deployed model multiple times, the attacker has obtained more knowledge from the leaked models, and can launch stronger adversarial attacks. Therefore,  the model owner requires a stronger notion of robustness than the one-time breach case, and a scalable method to extend the sequence of model deployment beyond the simple case of $\model_1$ and $\model_2$. 

In the following discussion, we consider a sequence of models $\model_1, ..., \model_i$ that has been deployed and breached, with $\model_i$ ($i \ge 2$) being the most recent model that is breached. Now the model owner needs to replace $\model_i$ with $\model_{i+1}$.

\para{Adversary Capabilities.} We consider a powerful attacker that has white-box access to all the breached models. With no knowledge of the replacement model $\model_{i+1}$, the attacker leverages transferability to attack $\model_{i+1}$.   The attacker is {capable} of constructing targeted adversarial examples using knowledge of $\model_1, ..., \model_i$, \ie generating $\tilde{x}=x+\delta_x$ such that, for at least one $j$ in $\{1, ..., i\}$, 
\[\model_j(\tilde{x})=l_t \neq \model_j(x), \;\; j\in \{1, ..., i\}\] where $d(x,\tilde{x})<\epsilon$.  \revision{We also consider ``highly cautious'' attackers who only apply $\tilde{x}$ to attack $\model_{i+1}$ if $\tilde{x}$ succeeds on {\em all} prior versions ($\model_1, ..., \model_i$).}

\para{Model Owner's Goal.} Besides having high performance and robust against any model that is previously breached, the model owner also needs 
\begin{packed_itemize} 
  \item a scalable approach to find the replacement model $\model_{i+1}$;
  \item $\model_{i+1}$ is robust against the strong attacker that has knowledge of all previous models $\model_{1}, ..., \model_{i}$. 
  \end{packed_itemize}
  
  \para{Robustness Measure.}  We define a new notion of robustness to characterize the strong attacker that has white-box access to all previous model versions.  This is measured by the attack transferability under attacks generated using an ensemble of $\model_1, ..., \model_i$, namely the {\bf {\em compound attack transferability}}:
  \[\trans(\{\model_j\}_{j=1}^{i} \rightarrow \model_{i+1}).\]
Correspondingly, the model owner needs a model versioning technique that produces a sequence of models $\model_1, ..., \model_i, \model_{i+1}$ to minimize the compound attack transferability at each model version. 
   
\subsection{Design Challenges and Initial Solutions}
\label{subsec:goal}
We identify two unique challenges facing the design of scalable and robust model versioning.

\begin{packed_itemize}
\item {\bf Versioning uncertainty} -- The model owner cannot foresee the exact number of model versions required beforehand, nor can they anticipate the timing of a potential breach of the deployed model. Yet the model owner must promptly identify a replacement model once a breach is detected.

\item {\bf Continuous expansion of attacker knowledge} --   With each occurrence of model leakage, the attacker gains more information about the models, making it hard to maintain  low compound attack transferability while simultaneously ensuring high performance on the primary task. 
  On the other hand, the model owner's lack of access to new training data hinders their ability to gather additional knowledge for constructing new models and defending against attacks.

\end{packed_itemize}

\label{subsec:challenge}

\para{Initial Solutions.}
Drawing upon existing literature, we identify several initial solutions to perform model versioning. We also explore their limitations, which motivate our work.   

\begin{packed_itemize}
\item \textbf{Varying model initialization} --  One might consider generating model variants by varying the model initialization \revision{before training}.  However, since these models are trained using the same dataset, they tend to converge with a high likelihood and share high attack transferabilities. 
Later, we validate this projection both theoretically and empirically.

\item \textbf{Varying training batch order} -- Similarly, one can produce model versions by injecting ``randomness'' to the order of training data.  Yet since the training data remains unchanged, these models also tend to converge with a high probability.

\item \textbf{Varying model architecture} --  One can employ distinct model architectures across model versions. Yet this method faces two critical limitations. First, the available model architectures for a given task are often limited, especially for high-performance models tailored to complex tasks. Second, the mere use of different architectures does not assure a reduction in transferability in a principled manner. 

\item \textbf{Generating model ensemble} -- Ensemble-based model optimization a(e.g., TRS~\cite{yang2021trs})  trains a fixed ensemble of models before deployment, aiming to improve ensemble robustness by reducing pairwise attack transferability among models in the ensemble. This approach is ill-suited for our specific problem due to the two distinct challenges outlined earlier: version uncertainty and continuous expansion of attack knowledge. The ensemble methods require knowledge of the exact number of model versions in advance, and once the ensemble is trained, there is no method for generating additional model versions. Furthermore, ensemble methods tend to be computationally expensive and primarily focus on reducing pairwise attack transferability among models. We confirm these via empirical experiments later on.

  \revision{\item \textbf{Detecting attack inputs} -- As discussed in \S\ref{sec:background},  recent work~\cite{shan2022post} develops a method to generate model variants by adding randomly generated images to model training data. Rather than reducing attack transferability, \cite{shan2022post} focuses on creating a separate input filter to detect transferability-based adversarial examples at run-time.  Consequently, as more model versions are leaked, the attack transferability remains high (\eg 90\%) while the input filter loses its effectiveness.  Our experiments also confirm this observation. 
  } 
\end{packed_itemize}
The above discussion shows that none of the existing solutions effectively address the challenges facing scalable and robust model versioning. 
In the following section, we introduce our proposed solution \revision{that significantly improves over existing solutions} by utilizing optimized hidden training to continuously generate robust model versions over time.

\section{Proposed Solution: Optimized Hidden Training}
\label{sec:hiddentrain}

We propose a new approach to model versioning, utilizing the novel concept of ``\revision{optimized} hidden training'' to continuously produce model versions that exhibit resilience against compound transferability attacks. Next, we present the intuition behind our solution, followed by the formal definition and optimization process for configuring hidden training.

\subsection{Design Intuition}
\vspace{-0.05in}
\para{Augmenting Model Training Using ``Hidden Data.''}
We propose to create distinct model versions by introducing \revision{carefully planned} variability into the model training data. At a broad level, our solution aligns with the concept of ``training data augmentation.'' However, what sets our approach apart is the automatic self-curation of additional training data for each model version, all without the necessity of acquiring new training data. In the following, we refer to this supplemental training data, which is unique to each model version, as ``hidden data'' because it is constructed internally with a focus on maintaining secrecy.

\para{Curating Hidden Data from a Single Feature Point.} Another distinctive contribution of our work is the novel concept of curating hidden data from a {\em single} feature point, which represents a parameterized, hidden distribution that is irrelevant to the classification task.   By varying the choice of this feature point,  we seek to naturally introduce variability to the loss surface and the “non-robust” features learned by the models. \revision{Later in \S\ref{sec:theory},  we analytically validate this design decision.} 

Specifically, we first establish the task feature space by training an original model $\modelori$ solely on the task training data $\trainori$. We denote the task feature space $\featurespaceori$. Next, we select a solitary feature point $\hiddenfeature_i$ within $\featurespaceori$ for the $i^{th}$ model version, and then create the associated data, which possesses a feature corresponding to $\hiddenfeature_i$ concerning $\modelori$.
Given that each model version $i$ has its own hidden data aligned with its unique $\hiddenfeature_i$, we can apply hidden training to generate multiple model versions that excel at the same task while exhibiting distinct attack characteristics.

\begin{figure}[t]
  \centering
 \includegraphics[width=0.23\linewidth, angle=270]{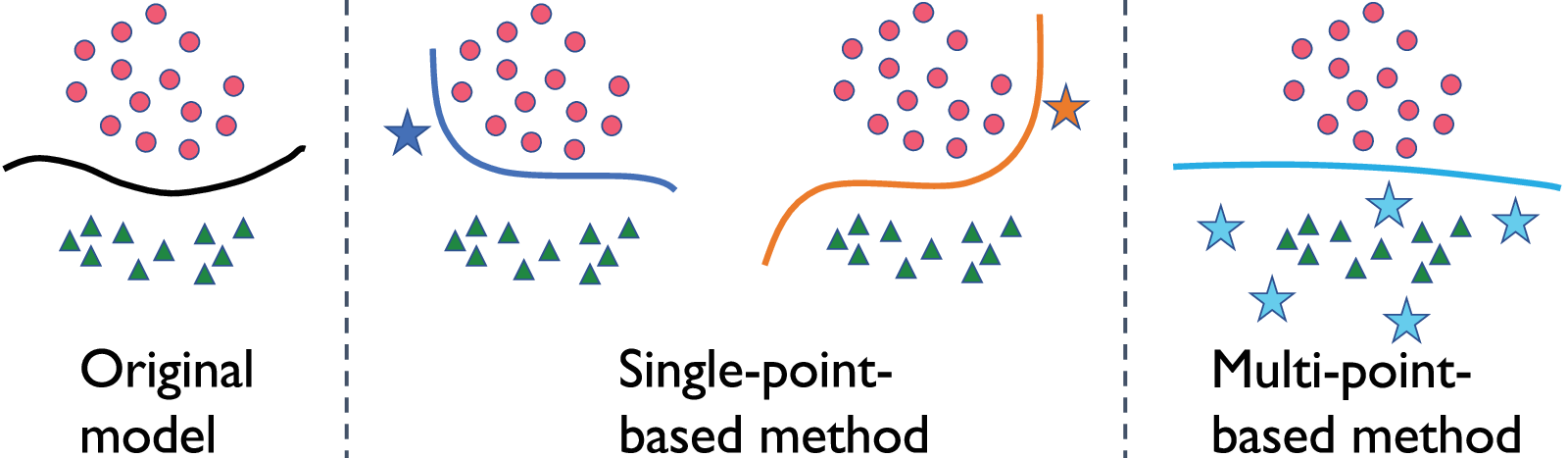}
 \caption{\em Impact of hidden training on models. By generating hidden data from a single feature point, one can  introduce unique modifications to model decision boundaries. But when generating hidden data from multiple feature points, both the extent and uniqueness of the modification decrease.} 
  \label{fig:hiddendata}
  \vspace{-0.15in}
\end{figure}

The question that naturally arises is why we opt for using a single feature point to construct the hidden data. We argue that this approach offers the following two significant advantages.  
\begin{packed_itemize} 
  \vspace{0.05in}
\item {\bf Adding unique ``distortions'' to decision boundary} -- Incorporating hidden data into the training process is expected to have an impact on the model's decision boundary within the feature space. Our strategy of concentrating all efforts on a single feature point is designed to efficiently ``adjust'' the decision boundary in the direction defined by that particular feature point. In contrast, if we were to generate hidden data from multiple feature points (similar to the conventional augmentation method), their effects on the decision boundary might counterbalance each other. This would not only lead to more subtle changes in the decision boundary but also diminish the uniqueness of those changes. In Figure~\ref{fig:hiddendata}, we present several illustrative examples where the single-point-based method introduces version-specific modifications to the decision boundaries, unlike the multi-point-based method.

\item {\bf Maintaining original task performance} -- Including non-task data during training may increase the training complexity.   When adding multiple feature points to a class, it is likely that the added training data disturbs the model performance on this class. Therefore, adding just a single feature point helps maintain high performance of the model without introducing high complexity to model training, making the training process as efficient as the original model.

\end{packed_itemize}

\subsection{Formal Definition}

We now formally define the formulation of hidden data and the process of hidden training to create new model versions.

\begin{definition}
  (\textbf{\revision{Parameterized} Hidden Data}) Let $\hiddenfeature$ be a point in $\featurespaceori$,  where $\featurespaceori$ represents the feature space of the model trained solely on the task training data $\trainori$. For $x \in \mathbb{R}^n$, we abuse the notation to let $\featurespaceori(x)$ denote the corresponding feature value of $x$ in the feature space $\featurespaceori$. Let $\hiddendata$ be data in the input space $\mathbb{R}^n$ such that $\forall x \in \hiddendata$, $d(\featurespaceori(x), \hiddenfeature)<\epsilon_h$, where $d(\cdot,\cdot)$ is a distance metric for $\featurespaceori$. To protect a specific class $l_t$ against adversarial examples, we label all $x \in \hiddendata$ as $l_t$, producing the following {\em hidden data} for this class: 
  \begin{equation*}
      \trainhidden = \{ (x, l_t) \;|\; x \in \hiddendata \}.
  \end{equation*}
\end{definition}

\begin{definition}
  (\textbf{Hidden Training})  Given the hidden dataset $\trainhidden$, the hidden training process is to train a new model from scratch utilizing the merged training data, $\trainori \cup \trainhidden$. This newly trained model is denoted as $\model_{\trainori \cup \trainhidden}$.
  \label{def:hidden_training} 
\end{definition}
\noindent In practice, one can implement the hidden training process via three consecutive steps:

\vspace{3pt} \noindent {\bf 1.}  Using Algorithm~\ref{alg:hidden_feature} to select \revision{a set of feature points $\{\hiddenfeature\}$ from the feature space $\featurespaceori$, one for each class to be protected.  To maintain high performance on the original task, they should avoid overlapping with the feature areas of  $\trainori$. }

\vspace{3pt} \noindent {\bf 2.}  Using Algorithm~\ref{alg:hidden_train} to construct the hidden data $\trainhidden$ based on $\{\hiddenfeature\}$ to ensure that the feature values of $\trainhidden$ closely match  each of $\hiddenfeature$.   Here the data curation employs a technique reminiscent of computing adversarial examples, wherein an input data is perturbed to adjust its feature value towards the target feature point $\hiddenfeature$. 

\vspace{3pt} \noindent {\bf 3.} Train a new model from scratch using $\trainori \cup \trainhidden$.

\vspace{2pt} Following this process, model owners can generate multiple model versions without acquiring new training data.  By  choosing $N$ different \revision{sets of} points in the task feature space, \revision{\ie $\{\hiddenfeature\}_1, ..., \{\hiddenfeature\}_N$}, they can produce $N$ different sets of hidden training data $\trainhidden_1, ..., \trainhidden_N$,  and correspondingly, $N$ different variants of the original model: 
$\model_{\trainori \cup \trainhidden_1}, ..., \model_{\trainori \cup \trainhidden_N}$.  For ease of notation, we hereby refer to those models as $\model_1$, ..., $\model_N$. This model versioning process is {\em efficient} and {\em scalable}, and there is no need to fix $N$ a priori.  Generating additional model versions beyond $N$ is achieved by  selecting new hidden feature points and repeating the hidden training process.  

\revision{To our knowledge, we are the first to present a formal definition of the robust model versioning problem where $\hiddenfeature$ is the key parameter. None of the previous work, including~\cite{shan2022post}, seek to understand impact of the choice of $\hiddenfeature$ on (compound) attack transferability against subsequent models.}

\vspace{-0.05in}
\subsection{Optimizing Hidden Training for Robustness}
Using hidden training, model owners can generate a sequence of model variants capable of performing the original classification task while resisting  transferability-based attacks. For a sequence of $N$ model versions, the level of robustness is contingent on the selection of $h_1, ..., h_N$ used to curate hidden training data for each model version. In this work, we consider two methods for selecting hidden features:  {\em random selection} and {\em greedy optimization}.  \revision{In \S\ref{sec:theory} and \S\ref{sec:system}}, we explore, both analytically and experimentally,  their effectiveness in producing a robust sequence of model versions.

\begin{packed_itemize}
\item {\bf Random selection} -- The simplest method is to randomly select a new feature point when generating a new model version~\cite{shan2022post}.  One might assume that the randomness in $h_1, ..., h_N$ would naturally introduce diversity among the resulting model versions. However, we demonstrate through both analytical and empirical results that the randomness in feature point selection does not necessarily diminish attack transferability. This becomes particularly true as the value of $N$ increases since the attacker accumulates more knowledge about the models with each successful model breach.

\item {\bf Greedy optimization} -- To resist compound transferability attacks, the model owner can leverage their own access and understanding of previous (and leaked) models to strategically choose the hidden feature points for the subsequent replacement model.  This selection is greedy as  the model owner does not know how many additional 
  model versions they need to generate a priori.  To recover from the loss of model version $i$, we pre-train the next version $i+1$, based on all previously breached versions $\model_1, ..., \model_i$. 
We defer the in-depth discussion of the algorithm to \S\ref{sec:system}, as it relies on the insights derived from the analytical study of hidden training to be presented in \S\ref{sec:theory}. 
\end{packed_itemize}

\section{Analytical Study of Hidden Training}
\label{sec:theory}
In this section, we present an analytical case study on model versioning using \revision{parameterized} hidden training. Our goal is to demonstrate the importance of carefully selecting hidden features when producing the sequence of robust model versions.   Our analysis focuses on binary classification tasks utilizing \revision{linear} Support Vector Machine (SVM) models. We examine how the choice of $\hiddenfeature$ impacts the model's decision boundary and explore different configurations of $\hiddenfeature$ to create a sequence of model versions that resists direct and compound transferability attacks over time.  \revision{We also discuss ways to generalize our analysis to more complex settings.}

\subsection{Preliminaries}

We consider a binary classification task with two classes over input space $\mathcal{X} \subseteq \mathbb{R}^2$. Let $\mathcal{X}_+$ and $\mathcal{X}_-$ denote the task data of class $+$ and class $-$,  respectively.  Let $\trainori_+$ and $\trainori_-$ denote the task training data of the two classes, respectively.  The detailed configuration of these datasets are shown in Table~\ref{tab:datadefine}. 
The task here is to train \revision{linear} SVM model versions to classify points from $\mathcal{X}$ to $\{+, -\}$, using the task training data $\trainori_+$, $\trainori_-$, and the chosen hidden data $\hiddendata$.

\para{Input Space = Feature Space.} We note that in the \revision{linear} SVM setting, the input space and the feature space are identical, \ie $\{\hiddenfeature\}=\hiddendata$.  Thus,  \revision{to streamline our discussion},  we directly use the chosen hidden feature $\{\hiddenfeature\}$ to represent $\hiddendata$. 
However, this statement does not hold in general, specifically for DNN models where the feature space is considerably different from the input space.

\para{Hidden Training in SVM.} An SVM model $\model: \mathbb{R}^2 \to \{+,-\}$ is a function that takes data from $\mathbb{R}^2$ as input and outputs the class label.
We assume that the class to be protected, \ie the class that will be targeted by the attacker, is class $+$. Hidden training involves choosing hidden data $\hiddenfeature$ and the target class $+$, and adding it to the training data set.

Our analysis of hidden training starts from the following theorem and the definition of attackable region.

\begin{theorem}
  \label{thm:hidden_determine}
  (\textbf{Hidden Training Determines SVM}).
  When $\hiddenfeature$ is in $\{ (x,y) \in \mathbb{R}^2 \; |\; |x| < \xcenter - 1, |y| \le y_{lim}\}$, $\hiddenfeature$ determines the decision boundary of the trained SVM model.
  \end{theorem}
\noindent \revision{As defined by Table~\ref{tab:datadefine},  $c$ is the x-axis center of the training data's feature cluster for $+$ class and $y_{lim}$ bounds the y-axis of the feature space.}  The proof of Theorem~\ref{thm:hidden_determine} can be found in Appendix~\ref{appendix:newSVM}.

This theorem shows that, in the two-dimensional space, the optimal linear SVM classifier is the one that bisects the shortest connection between the convex hulls of the training data in the classes~\cite{scholkopf2002learning}.  
When $\hiddenfeature$ is not in $\trainori_+$, $\hiddenfeature$ determines the convex hull of $\trainori_+ \cup \trainhidden$. A properly chosen $\hiddenfeature$ can also determine the SVM classifier when it changes the shortest connection between the convex hulls. Thus, Theorem~\ref{thm:hidden_determine} characterizes the region in $\mathbb{R}^2$ where the SVM is determined by the hidden data added to class $+$.

\para{Attackable Region.}  Given a model version $\model_i$, we define its attackable region to identify the subspace of $\mathbb{R}^2$ where any adversarial attacks targeting class $+$ can appear. This region contains all possible attacks generated by the strongest adversary, one who operates without any limitations on perturbation budgets or computational resources.  We use $\attackr_i$ to denote the attackable region of $\model_i$.

\begin{definition}
  ({\bf Attackable Region})  For an SVM model $\model_i$, we define its attackable region for the target class $+$ as  \[\attackr_i = \{(x,y) \in \mathcal{X}_- \; | \; \model_i(x,y) = +\}.\]
\end{definition}

\noindent The attackable region of model $\model_i$ contains all input data that should be categorized as class $-$ but wrongly classified as class $+$ by the model. That is, any possible attack directed towards class $+$ that an attacker can construct using only $\model_i$ falls within $\attackr_i$. Beyond $\attackr_i$, the data is either correctly classified by the model or bears a ground truth label of $+$.

 \begin{table}[t]
	\centering
	\begin{tabular}{ c | p{6.2cm}} 
		\hline
		  \begin{tabular}[c]{@{}l@{}}Task data\\ ($\mathcal{X}_+$, $\mathcal{X}_-$)\end{tabular}
                         & Data in $\mathcal{X}_+$ and in $\mathcal{X}_-$ are symmetrically distributed about the y-axis. For an input $(x, y) \in \mathcal{X}$, we have $(x, y) \in \mathcal{X}_+$ if $x \ge \delta$ or $-\delta < x < 0$ where $\delta > 0$. Otherwise, $(x, y) \in \mathcal{X}_-$.
                           Moreover, we bound the space such that for all $(x, y) \in \mathcal{X}, |y| \le \ylim$. 
Figure~\ref{fig:SVM} illustrates the setup.
            In this setting, no SVM model achieves perfect accuracy on $\mathcal{X}$. However, there exists SVM models that can linearly separate our training data, defined as below. \\ \hline
			
			\begin{tabular}[c]{@{}l@{}}Task training data\\ ($\trainori_+$, $\trainori_-$) \end{tabular}
             &  Data in $\trainori_+$ is uniformly distributed in a unit circle centered at $(\xcenter,0)$ and $\trainori_-$ in a unit circle centered at $(-\xcenter,0)$ where $c \gg \delta$. We use $\trainori$ to denote the clean training data where $\trainori = \trainori_+ \cup \trainori_-$. Therefore, $\mathcal{X}$ is not linearly separable but $\trainori$ is. \\
		\hline
	\end{tabular}
	\caption{\em Configuration of task data and training data.}
	\vspace{-0.15in}
	\label{tab:datadefine}
\end{table}

\subsection{Impact of Hidden Data Choice: One-time Breach}
\label{subsec:direct_theory}

Next, we study how the choice of hidden data $h$ affects the robustness of model versions, starting from the case of one-time breach. For this simple case, the robustness is measured by the directional attack transferability from $\model_1$ to $\model_2$.   Our analysis focuses on modeling the directional attack transferability using the area of the attackable region.   For two models $\model_1$ and $\model_2$ trained with different hidden data, all transferable attacks exist in the intersection of $\attackr_1$ and $\attackr_2$. 

\begin{definition}
({\bf Directional Attack Transferability}) 
Let $\areafunc: \mathbb{R}^2 \to \mathbb{R}$ be a function that computes area of a subspace in $\mathbb{R}^2$. The directional attack transferability from $\model_1$ to $\model_2$, denoted as $\trans (\model_1\rightarrow \model_2)$, is computed as:
\begin{equation*}
\trans (\model_1 \rightarrow \model_2) =\frac{\areafunc(\attackr_1 \cap \attackr_2)}{\areafunc(\attackr_1)}
\end{equation*}
\label{def:directional_attack_transfer}
\end{definition} 
\vspace{-0.2in}
\noindent When the two models share no attackable region (\ie $\attackr_1 \cap \attackr_2 = \emptyset$), the directional attack transferability becomes zero. The following theorem defines a condition on the choice of $h_1$ and $h_2$ to meet such condition.

\begin{figure}[t]
  \centering
  \vspace{-0.2in}
    \includegraphics[width=0.65\linewidth, angle =270]{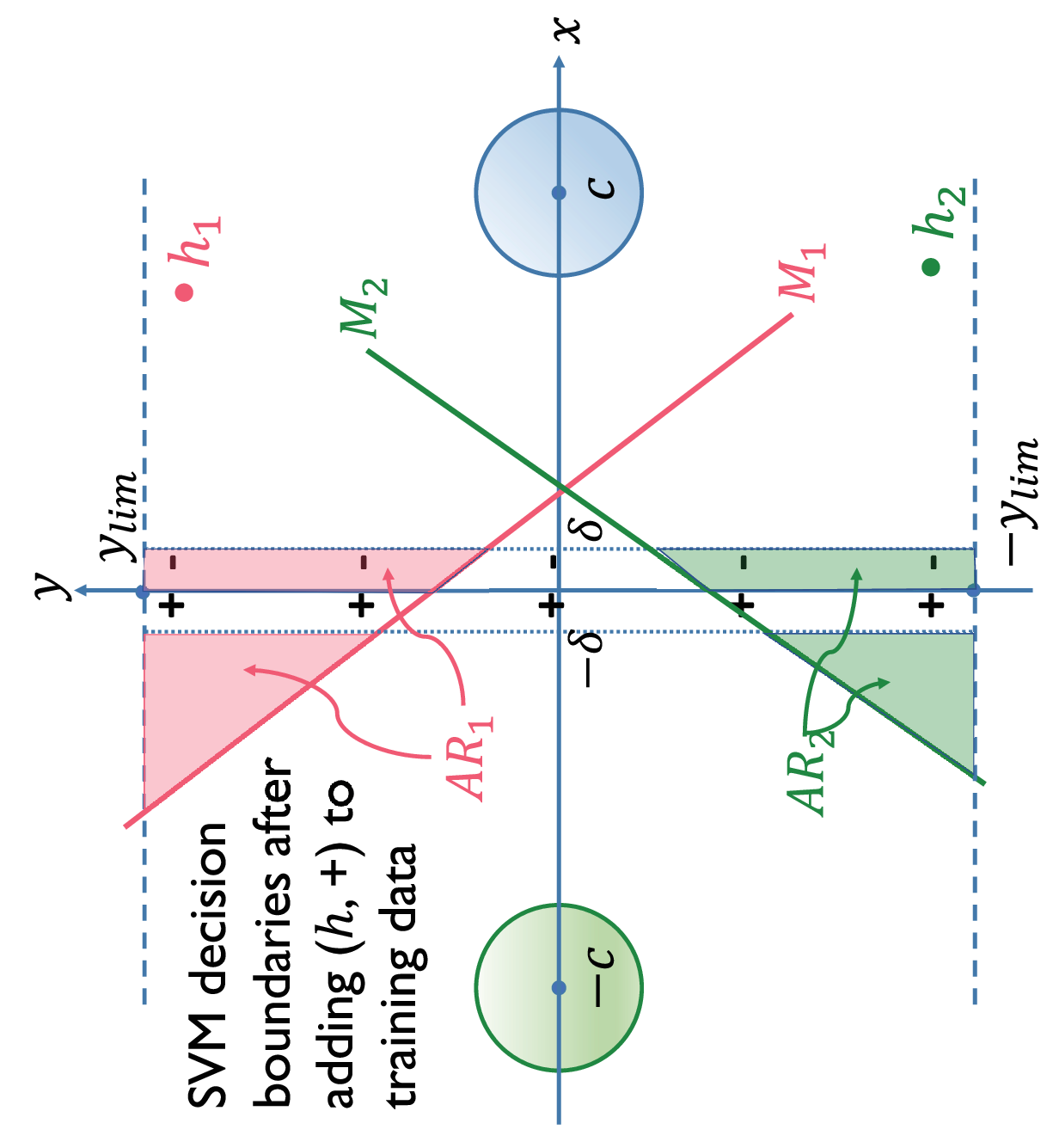}
    \caption{\em Illustration of SVM decision boundaries after hidden training using $(\hiddenfeature_1, +)$ and $(\hiddenfeature_2, +)$, respectively. The two resulting models $\model_1$ and $\model_2$ share zero attack transferability because their attackable regions do not overlap.}
    \label{fig:SVM}
     \vspace{-0.15in}
\end{figure}

\begin{theorem}
\label{thm:no_transfer}
(\textbf{Nullifying Directional Attack Transferability})
For two model versions $\model_1$ and $\model_2$ produced from hidden training, we have $\trans (\model_1\rightarrow \model_2) = 0$ when decision boundaries of $\model_1$ and $\model_2$\\
(1) have opposite signs of slope, and;\\
(2) intersect at $(x_I, y_I)$ with $x_I > \delta$.
%
\end{theorem}

\noindent The proof of Theorem~\ref{thm:no_transfer} is in Appendix~\ref{appendix:no_transfer}. In the proof, we also illustrate how to find qualified $\hiddenfeature_1$ and $\hiddenfeature_2$ to achieve zero transferability.  \revision{We first exclude any linear separator that is not achievable by adding hidden data $\{\hiddenfeature, +\}$ to $\trainori$. Next, given an achievable} linear separator, we reconstruct the coordinates of $\hiddenfeature$ by finding the connection between the two convex hulls bisecting by the separator. 
After identifying an achievable linear separator with hidden data $\hiddenfeature_1$ that intersects the x-axis at a point $>\delta$, we  select  $\hiddenfeature_2$ as the coordinate symmetrical to $\hiddenfeature_1$ along the x-axis, and use it to  train $\model_2$.  
Figure~\ref{fig:SVM} illustrates a specific scenario where the two models have no overlap in their attackable regions, resulting in zero attack transferability.

\revision{\para{Additional Observations.}  Theorem~\ref{thm:no_transfer} leads to three additional findings, which later inform our approach in designing practical algorithms for selecting hidden features (\S\ref{sec:system}).}
\begin{packed_itemize} 

\item {\bf Random $h$ values can rarely produce SVMs with low attack transferability} --  Theorem~\ref{thm:no_transfer} demonstrates that $\hiddenfeature_1$ and $\hiddenfeature_2$ must be located on opposite sides of the x-axis to minimize directional attack transferability. Yet if  $\hiddenfeature_1$ and $\hiddenfeature_2$ are randomly selected, there is a 50\% chance that both fall on the same side of the x-axis, resulting in a high  attack transferability between the two model versions.

\item {\bf Varying training parameters does not lower attack transferability} -- In the SVM setting, adjusting model training parameters such as initialization and training batch order has no impact on the trained model. Therefore, in practice, such variations are unlikely to result in model versions with reduced attack transferability.

\item {\bf Separation between hidden data and task data} -- Effective hidden data is often situated at a considerable distance from the training data in the feature space. \arevision{Thus, we} look for candidates of hidden features in outliers within each class.

\end{packed_itemize}

\subsection{Impact of Hidden Data Choice: Repeated Breaches}
\label{subsec:generalN}

We now consider a sequence of $N$ ($N>2$) model versions ($\model_1, ..., \model_\totalversions$).   Within this sequence, after launching $\model_{i}$, an attacker,  who has white-box access to $\model_1, ..., \model_{i-1}$ but not $\model_{i}$, can attack $\model_{i}$ using a combined  knowledge of all previous model versions, \ie launching a compound transferability attack.  We first establish a theoretical model for the compound attack transferability in the SVM case, utilizing the attackable region concept.  All potential compound attacks are found in the union of $\attackr_1, \ldots, \attackr_{i-1}$, and the successful ones against $\model_{i}$ are situated in the intersection of $\attackr_{i}$ and the aforementioned union.

\begin{definition}
({\bf Compound Attack Transferability}) 
The compound attack transferability from $\model_1, ..., \model_{i-1}$ to $\model_i$, denoted as $\trans(\{\model_j\}_{j=1}^{i-1} \rightarrow \model_i)$, is computed as:
\begin{equation*}
  \trans(\{\model_j\}_{j=1}^{i-1}
  \rightarrow \model_i) = \frac{\areafunc(\attackr_i \cap (\cup_{j=1}^{i-1} \attackr_j))}{\areafunc(\cup_{j=1}^{i-1} \attackr_j)}
\end{equation*}
\label{def:compound_attack_transfer}
\end{definition}
\vspace{-0.2in}
\noindent Definition~\ref{def:compound_attack_transfer} allows us to compute the attack transferability against any model version within  a sequence of model versions. In this model, we characterize the strongest attacker that can utilize all attackable regions of previous leaked versions. That is, any attack generated from an ensemble of the previous versions is included by the union of attackable regions.

Next, we show that one can employ greedy search to construct a sequence of model versions, where the compound attack transferability within the sequence is upper bounded. 

\begin{theorem}
\label{thm:sequence}
(\textbf{Greedy Search Can Upper Bound Compound Transferability})
Using greedy search, we can construct a sequence of model versions $\model_1, ..., \model_{\totalversions}$ such that, 
\begin{equation*}
 \max_{i, i\leq N} \trans(\{\model_j\}_{j=1}^{i-1} \rightarrow \model_i) \le \alpha_\totalversions
\end{equation*}  
  where $\alpha_\totalversions$ increases with $\totalversions$.
\end{theorem}	\vspace{-0.05in}
\noindent The detailed proof is in Appendix~\ref{appendix:compound}. Here we briefly sketch the greedy search process.  First, we construct $\model_1$ and $\model_2$,  leveraging Theorem~\ref{thm:no_transfer} to identify $h_1$ and $h_2$ such that $\trans (\model_1 \rightarrow \model_2) = 0$, and their combined attackable region $\attackr_1 \cup \attackr_2$ is sufficiently large to 
minimize overlapping portion with subsequent model versions.
In our design,  $\cup_{j=1}^{i} \attackr_j = \attackr_1 \cup \attackr_2$ for any $i \ge 2$ and the decision boundaries of $\model_1$ and $\model_2$  are approximately orthogonal.
 Next, we select $h_3$ so that the decision boundary of $\model_3$ is in parallel to that of $\model_1$ but shifted to create a much  smaller attackable region.  Since $\attackr_3 \cap (\attackr_1 \cup \attackr_2)$ is small,  the compound attack transferability $\trans(\{\model_1, \model_2\}
\rightarrow \model_3)$ is low.  Similarly, we select $h_4$ so that $\model_4$ has a decision boundary parallel to $\model_2$, while $\attackr_4$ is much smaller than  $\attackr_2$. Also, the decision boundaries of $\model_3$ and $\model_4$ are approximately orthogonal.  Following this alternating strategy, we can progressively deploy the subsequent model versions, yet each time we gradually {\em increase} the attackable region.  As a result, the decision boundaries of $\model_i$ and $\model_{i+2}$ are parallel while those of $\model_i$ and $\model_{i+1}$ are approximately orthogonal.

Our strategy applies greedy optimization to find the best model version for the current version $i$, and does not assume the knowledge of $N$ when choosing $\model_i$. The maximum compound transferability $\max_{i, i\leq N} \trans(\{\model_j\}_{j=1}^{i-1} \rightarrow \model_i)$ is zero at $N=2$ and gradually increases with $N$, because the intersection of attackable regions with the previous versions increases with $N$.  As an example, Table~\ref{tab:ATofN} shows the upper bound $\alpha_\totalversions$ value as a function of $N$, under a specific configuration of data parameters (\ie those defined by Table~\ref{tab:datadefine}).  Here $\alpha_\totalversions$ increases gracefully with $N$, demonstrating the robustness of model versioning  using hidden training.

\begin{table}[t]
		\centering
			{
				\begin{tabular}{l|l l l l}
					\hline
					$\totalversions$                               & $ 2 $ & $4$ & $ 6$ & $8$ \\ \hline
					$\alpha_\totalversions$ & 0   & 0.17   & 0.37 & 0.4     \\ \hline
				\end{tabular}
			}
			\caption{\em A realization of the model sequence and the corresponding upper bound on the compound attack transferability.  The upper bound  increases with length of model sequence $\totalversions$.}
			\vspace{-0.2in}
			\label{tab:ATofN}
	\end{table}

 \revision{\subsection{Generalizing the Analysis}
   While our theoretical analysis targets two-dimension binary classification settings utilizing linear SVMs,  it offers insights that can be applied to construct robust model versions for more complex classification tasks.   Next we discuss directions in which our analysis can be expanded to those settings.


\para{High Dimensional Settings.} Extending our proof to binary classification in higher dimensions is relatively straightforward. In a $d$-dimensional space,  the corresponding hidden features lie in a $(d-2)$-dimensional space, \eg  in 3-D settings, the \arevision{effective} hidden features are confined to a line. We can apply similar methods used by the proofs of Theorems~\ref{thm:hidden_determine} and~\ref{thm:no_transfer} to find the optimal hidden features.


\para{Multi-class Settings.} Extending our analysis to multi-class SVM is much more challenging. There exist two main methods to reason about multi-class SVMs, One versus One (OVO) and One versus Rest (OVR).  For OVO, our proof naturally extends when considering protecting one class ($l_1$) against another class ($l_2$), by deriving the corresponding attackable regions. If the goal is to protect $l_1$ against all other classes ($l_2$, $l_3$, etc.), the attackable region will be the union of all the attackable regions against $l_1$.  However, since this region is hard to quantify analytically, it is hard to find a closed form derivation of the optimal $h$.
For OVR, our proof directly applies if we can assume that all data belonging to the rest of the classes lies on one side of the ($y=0$) line, \ie  we consolidate them into a single class. If the other classes are not linearly separable from $l_1$, then we need to employ soft-SVMs, which do not have a closed form solution.  

\para{Complex Feature Extractors.} Our analysis can be extended to models that employ complex feature extractors such as DNNs and Kernel SVMs.  For this we formulate the theoretical problem where the linear SVMs (used in our analysis) operate on the feature space instead of the input space. Thus, our analysis directly models the hidden features ($h$) that can reduce attack transferability.  Here we need to make two assumptions. First, there exists an inversion process to create data in the input space that realizes the chosen hidden features. In fact, our empirical algorithm for DNN models does use such an inversion process in Algorithm~\ref{alg:hidden_train}.  Second, the kernel or feature mapping of the input space must satisfy the properties described in Table~\ref{tab:datadefine}. The latter may not hold in practice exactly, but our empirical results show that the same insights do apply.}

\section{Generating DNN Model Versions}
\label{sec:system}
\revision{  Our analysis in \S\ref{sec:theory} 
  establishes the theoretical 
  foundation for the task of model versioning, but also suggests
  design principles for developing practical versioning algorithms
  for DNN models. We now present these principles and our
  detailed algorithm design.

\subsection{Design Principles}
\label{subsec:principles}
Our theoretical analysis produces three key guidelines for selecting hidden features. 
 
\begin{packed_itemize}
\item Theorem~\ref{thm:hidden_determine} shows that  a single
  $\hiddenfeature$ (per protected class)  is not only adequate but also serves as a
preferred optimization factor for perturbing feature space
and altering attack landscape. 
We follow this format to choose $h$ as the anchor for generating hidden data in the input domain.

\item Theorem~\ref{thm:no_transfer}
   shows that effective hidden data exists outside the convex hull of
   the training data and is relatively distant from the training data.
   Thus, in our practical algorithm,  we set the requirements that
   $\hiddenfeature$ does not overlap with features of the task training data and must keep a minimum distance to the center of training data in the feature space. 

\item Both Theorem~\ref{thm:no_transfer} and~\ref{thm:sequence} show that optimizing the choices of
  $\hiddenfeature$ can effectively reduce attack transferability
  compared to random selection.  Furthermore, the 
optimal locations of hidden features do not solely depend 
on their distance to the original feature clusters but  exhibit
a complex geometric relationship.  This leads us to design a
greedy-search based optimization method for locating  the right $h$ values.

  \end{packed_itemize} 
}

\subsection{Detailed Algorithm Design} 
We now present the
detailed algorithm for DNN model versioning. Previously, Definition~\ref{def:hidden_training} already outlines the three sequential steps for creating a model
version after selecting a feature point $h$ (per protected
class). \revision{Thus, our discussion below focuses on how to select the set of feature points in
$\featurespaceori$, one set for each model version.  To streamline our
presentation, we assume a single class is being protected, \ie one $h$
value per model version. The complete
process for protecting one or more classes is listed in Appendix~\ref{appendix:hiddentrain}}.

\para{Choosing $h$.}
Conceptually, one might expect that opting for $h_i$ values displaying greater
distance from $h_1, ..., h_{i-1}$ 
in the feature space would produce a more distinct model version
$\model_i$.  Similarly, choosing $h_i$ far from the
feature clusters of the target class would also improve the model robustness.  However, our empirical experiments show that there is no meaningful correlation between
such distances\footnote{We used three different distance metrics:
  averaged pairwise $\ell_2$ distances, averaged pairwise cosine
  distances and the Earth Mover's distance (also known as
  1-Wasserstein). The results are consistent across all three
  metrics.}, for either the attack transferability or the normal classification
accuracy. These findings align with the third design principle in
\S\ref{subsec:principles}.

Driven by these findings, we develop a greedy search approach for
selecting hidden features from a predefined candidate pool. To
build this candidate pool, we first identify a few ``edge'' points
within each class's feature cluster and rescale the 
feature vectors of these edge points to move them farther away from the
class cluster.  This ensures a sufficient gap between the hidden
data and any task-specific data,  \revision{following the second design
principle in \S\ref{subsec:principles}.} Next, we train models using these feature point candidates, producing a pool of candidate
models.  Finally, we choose the sequence of model versions, one model
at a time,  from this
model pool.   Each time, we first generate instances of compound transferability attacks based on
Projected Gradient Descent (PGD), leveraging white-box access to
all previous model versions. Then, we select, from the pool,  the
model candidate that has 
the lowest attack success rate as the subsequent model version in the
sequence. 

\para{Online vs. Offline Model Generation.}  Our algorithm for generating model versions adopts a greedy approach,
\ie generating models one by one.  Yet it presents the model owner
with two
options: (i) waiting to train a
replacement model only after detecting a breach in the current
version, and (ii) pretraining  multiple model versions in advance. The
second option takes more computation power, but offers the key
advantage of immediate model replacement.  This is because once a model is known to be
breached, leaving it active is clearly undesirable. Similarly, taking
the system offline while training a new model version is also
undesirable. Swapping in a pretrained model version minimizes security
risk and downtime, and is particularly important for large models that
take a long time to train.   Finally, for our algorithm, pretraining
(\eg $N$) models {\bf does not} restrict the model owner from
generating additional model versions beyond $N$.

\para{Curating Hidden Data.}  As discussed in
Definition~\ref{def:hidden_training}, we curate hidden data by
choosing a set of initial images and perturbing them 
using the technique proposed by~\cite{shafahi2018poison} so that their features
in the task feature space $\featurespaceori$ closely match $h$.   The choice of initial images is flexible. For our
implementation, we use a well-trained GAN
model~\cite{goodfellow2014generative, karras2017progressive} to
generate initial images. Appendix~\ref{appendix:add_result}
shows samples of hidden data (\ie perturbed GAN images) produced by  our
experiments.  We emphasize here that the key property of
the hidden data is not what it looks like in the input space, but
rather how close its representations are to the chosen feature point in the feature space.

\section{Empirical Study} 
\label{sec:exp}

In this section, we evaluate our \arevision{hidden training} method on
DNN models, using three image classification tasks. We also compare our method to \arevision{alternative methods to version models.}

\subsection{Experiment Setup}
\label{subsec:exp_setup}
\vspace{-2pt}
\para{Datasets and Architectures.} 
We consider  three image classification
tasks.
\begin{packed_itemize}
  \item {\bf $\cifar$}~\cite{krizhevsky2009cifar} is widely used to evaluate adversarial attacks
    and defenses. The task is to classify $10$ objects with $50,000$
    training images and $10,000$ testing images. The default model
    architecture is ResNet-18~\cite{he2016deep}, and the number of
    classes is $10$.  We also experiment with VGG-16~\cite{simonyan2014very}. 
  \item {\bf $\skincancer$}~\cite{tschandl2018ham10000} consists of $10,000$
    dermatoscopic images collected over $20$ years. The task is to 
    recognize $7$ types of skin cancer with $8,912$ training images
    and $1,103$ testing images. The
    model architecture is Densenet-121~\cite{huang2017densely}. 
  \item {\bf $\ytface$}~\cite{youtubeface} contains face images of $1,283$
    people ($587,137$ training images and $6,4150$ testing
    images). The task is to perform face recognition. The
    model architecture is ResNet-50~\cite{he2016deep} and the number
    of classes is $1283$. 
\end{packed_itemize}

\para{Attack Configurations.}  We consider three well-known
white-box adversarial attacks: Projected Gradient Descent
(PGD)~\cite{kurakin2018adversarial}, Carlini \& Wagner attack
(CW)~\cite{carlini2017towards}, and Elastic-net Attack
(EAD)~\cite{chen2018ead}. We use them to build both directional and compound
transferability attacks.  By default, we show results of PGD since it
leads to the highest attack transferability. For PGD attacks,  the
  $L_\infty$ perturbation budget is $0.03$ for $\cifar$,  $0.05$ for
  $\skincancer$, and $0.25$ for $\ytface$. 

  \arevision{In our experiments, 
    the attacker only submits an attack instance against the target 
    model if it has succeeded on prior model(s). Thus, the attack success
    rate equals the attack transferability defined in
    \S\ref{sec:problem}.  Specifically,  to launch {\em directional} transferability attacks
    against $\model_i$, the attacker has white-box access to only one
    prior model version $\model_j$, runs a standard white-box
    attack to create 1000 attack instances that succeed on
    $\model_j$,  and applies them to $\model_i$.   For \emph{compound}
    transferability attacks against $\model_i$, we use the ensemble
    attack from Tram\`er \emph{et. al.}~\cite{tramer2017ensemble} that
    leverages white-box knowledge of all previous model versions to
    create 1000 attack instances, ensuring that they succeed on at
    least one previous model version.  We also consider a ``cautious''
    attacker who only launches an attack instance against $\model_i$
    if it succeeds on {\em all} prior model versions. For all 
    cases, we report the attack transferability (and thus the attack success
    rate) as the fraction of attack instances launched against the
    target model $\model_i$ that actually succeed on $\model_i$.}

\para{Training Configuration.}  In hidden training, by default, we use hidden data
that is $20$\% of the original training data of the target class. Later in \S\ref{subsec:ablation} we
show that varying this ratio between $10$\% and $30$\% leads to the 
same conclusion. We set the model pool size to
$50$. \revision{Additional details of training and attack configurations are in Table~\ref{tab:config} in Appendix~\ref{appendix:config}.}

  \begin{table}[t]
	\centering
	\begin{tabular}{ l|l|l} 
		\hline
		& Original & Hidden Training  \\
		\hline
		$\cifar$ & $92.1 \%$ & $91.4 \pm 0.2 \%$  \\ 
  		$\skincancer$ & $88.7 \%$ & $90.5 \pm 0.5 \%$ \\ 
  		$\ytface$ & $98.9 \%$ & $98.6 \pm 0.4 \%$ \\         
		\hline
	\end{tabular}
	\caption{\em Performance of original and hidden trained models
        on clean inputs.}
	\vspace{-0.1in}
	\label{tab:clean_acc}
\end{table}

\subsection{Performance of Hidden Training}
\label{subsec:exp_hidden}
We evaluate the effectiveness of hidden training by 
{\em normal classification accuracy}, {\em scalability}, and {\em
  robustness to attacks}.

\para{Normal Classification Accuracy.} One of the
objectives of hidden training is to develop models whose normal model
accuracy is on par with the original model (without any hidden
training). To verify this, for each classification task, we generate 50 model
versions using hidden training and an original model without hidden
training.  In Table~\ref{tab:clean_acc}, we report the
normal classification accuracy of all 50 model versions for each task,
in terms of
mean and standard deviation,  along with that of 
the original model.  We see that the models trained using hidden
training achieve a normal accuracy comparable to that of the original
model.  

Interestingly, for $\skincancer$,  hidden training achieves
a higher mean normal accuracy than the original model.   This is
likely because  $\skincancer$'s limited training sample size and large
sparsity of the training images. In this case, the hidden data
injected during training essentially functions as a type of data
augmentation, which enhances model generalizability.

\para{Richness of Replacement Models.}  When a deployed model is breached  by attackers, the model
owner quickly recovers by deploying another model version 
that shares low attack transferability from the breached one(s).  
It is important to ensure that the set of qualified replacement models
is rich enough so that the adversary cannot easily enumerate through
the space to predict/construct the replacement model version. 

We evaluate the richness of replacement models as follows.  Given
the $50$ model versions generated by hidden training, we pick a model
whose average attack transferability to the other $49$ models is low,
and set it as the breached model to be replaced.  We then examine the
rest $49$ models and study their directional attack transferability {\em from}  the breached model.   Table~\ref{tab:replacement} presents the results for $\cifar$,  where for $25$ of the $49$ models, the
directional attack transferability from the breached model is less than $0.2$.

\begin{table}[t]
	\centering
	\begin{tabular}{ l|l|l|l} 
		\hline
		$\trans = \trans(\model_1 \rightarrow \model_i)$ & $\trans < 0.2$ & $\trans < 0.3$ & $\trans < 0.4$ \\
		\hline 
		\# of qualified model versions & 25              & 45              & 49              \\ 
		\hline
	\end{tabular}
	\caption{\em Richness of replacement model pool.}
	\vspace{-0.2in}
	\label{tab:replacement}
\end{table}

  \begin{figure*}[h]
    \centering
    \begin{minipage}{0.3\textwidth}
		\centering
		\includegraphics[width=\textwidth]{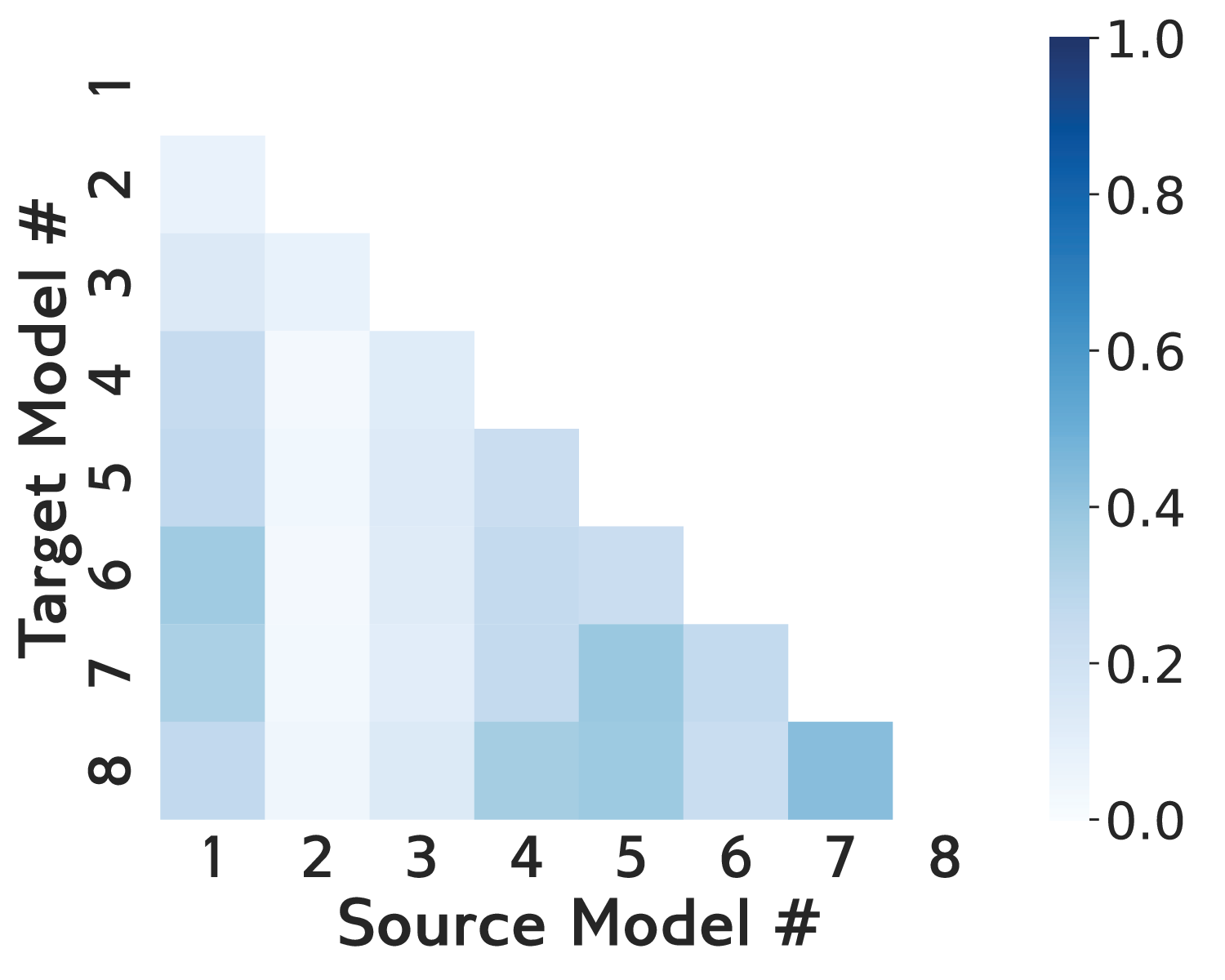}
		\caption{\em Directional attack transferability within
                  the model sequence ($\cifar$, PGD). Mean=0.19.}
		\label{fig:cifar_pair}
	\end{minipage}
    \hspace{7pt}
    \centering
    \begin{minipage}{0.3\textwidth}
        \centering
        \includegraphics[width=\textwidth]{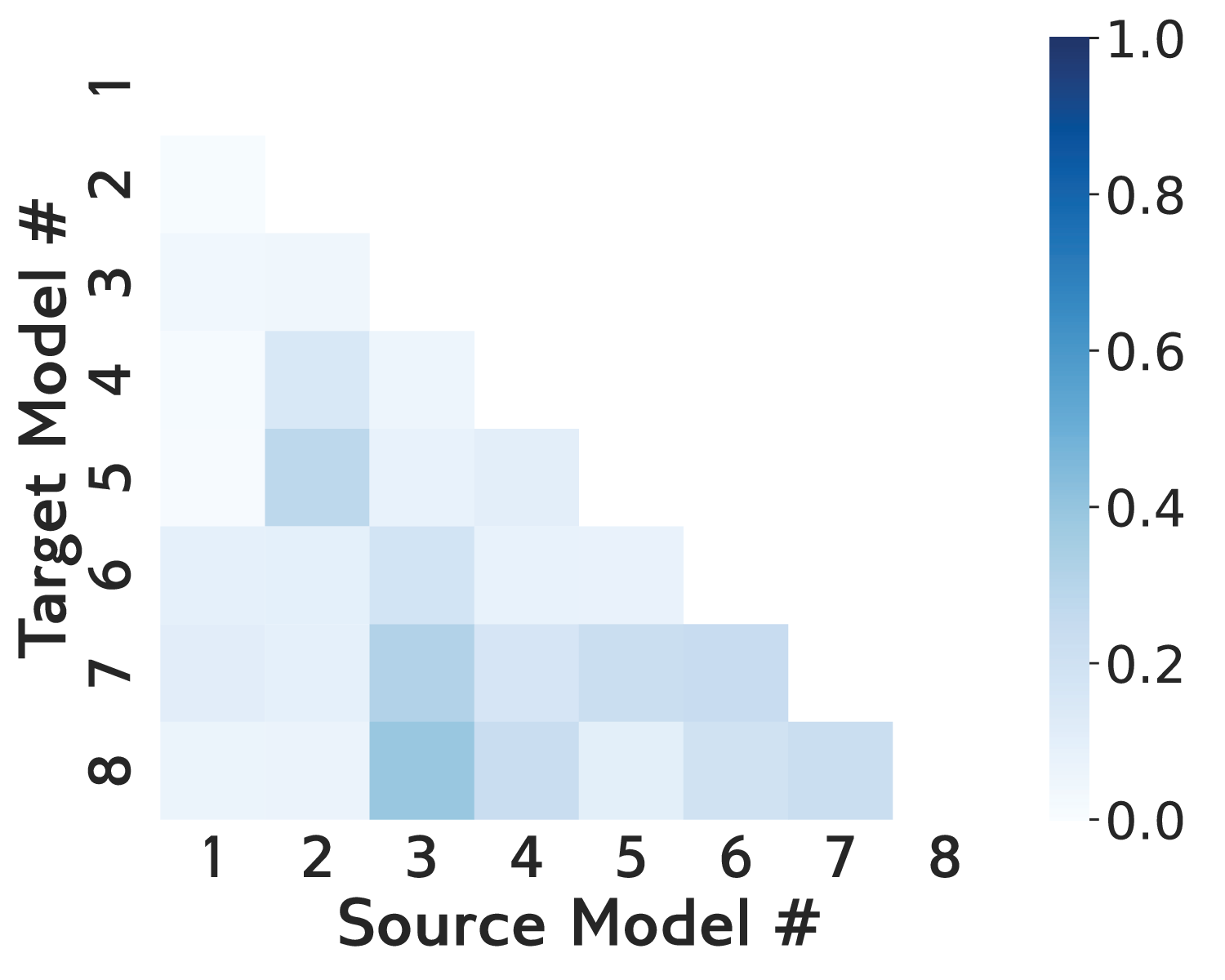}
        \caption{\em Directional attack transferability within the model
          sequence ($\skincancer$, PGD). Mean=0.13.}
        \label{fig:skincancer_pair}
    \end{minipage}
    \hspace{7pt}
    \centering
    \begin{minipage}{0.3\textwidth}
		\centering
		\includegraphics[width=\textwidth]{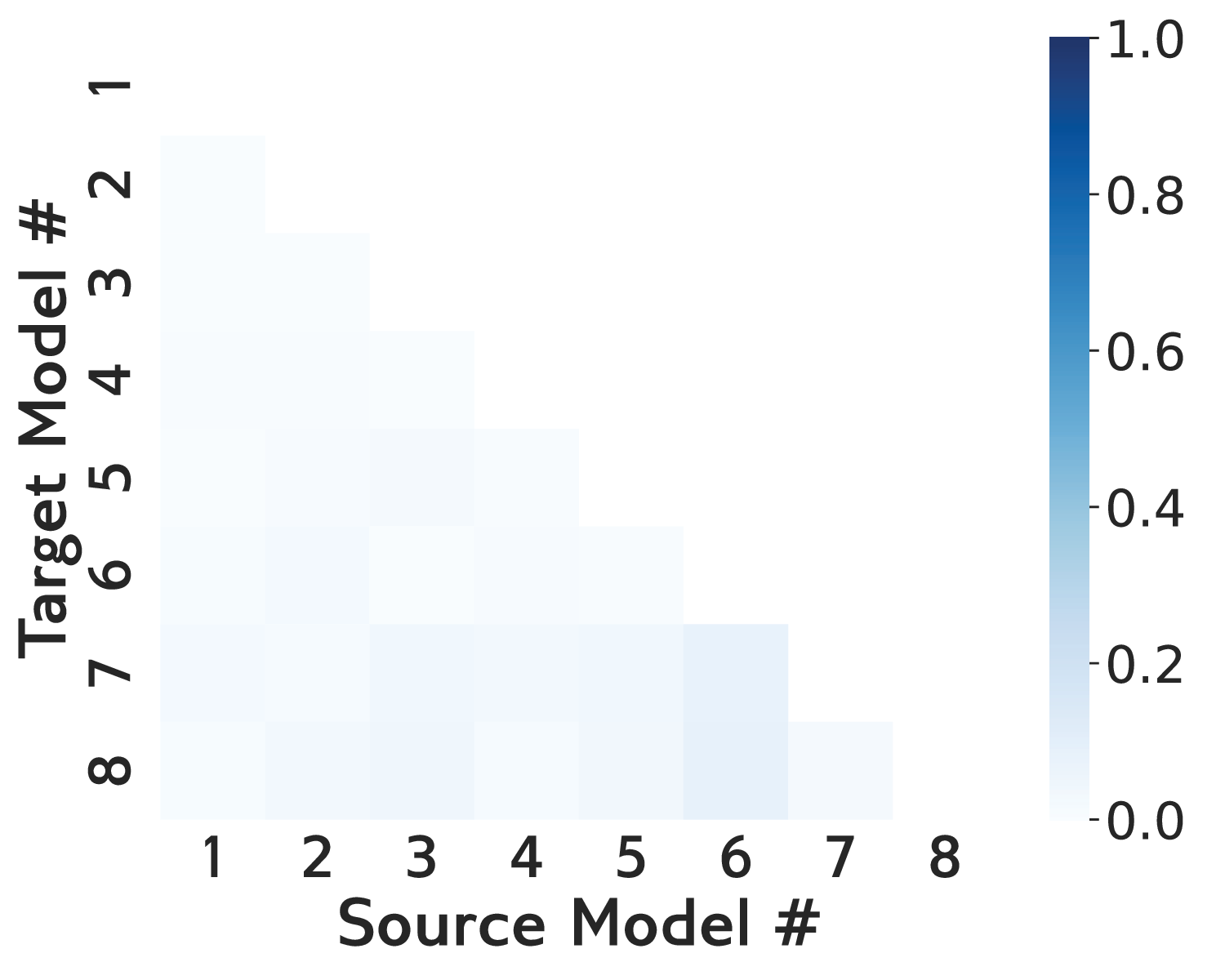}
		\caption{\em Directional attack transferability within the model
          sequence ($\ytface$, PGD). Mean=0.02.}
		\label{fig:ytface_pair}
	\end{minipage}	
    \hfill
    \vspace{-0.1in}
\end{figure*}

\begin{figure*}[t]
	\centering
	\begin{minipage}[t]{0.31\textwidth}
    \includegraphics[width=\textwidth]{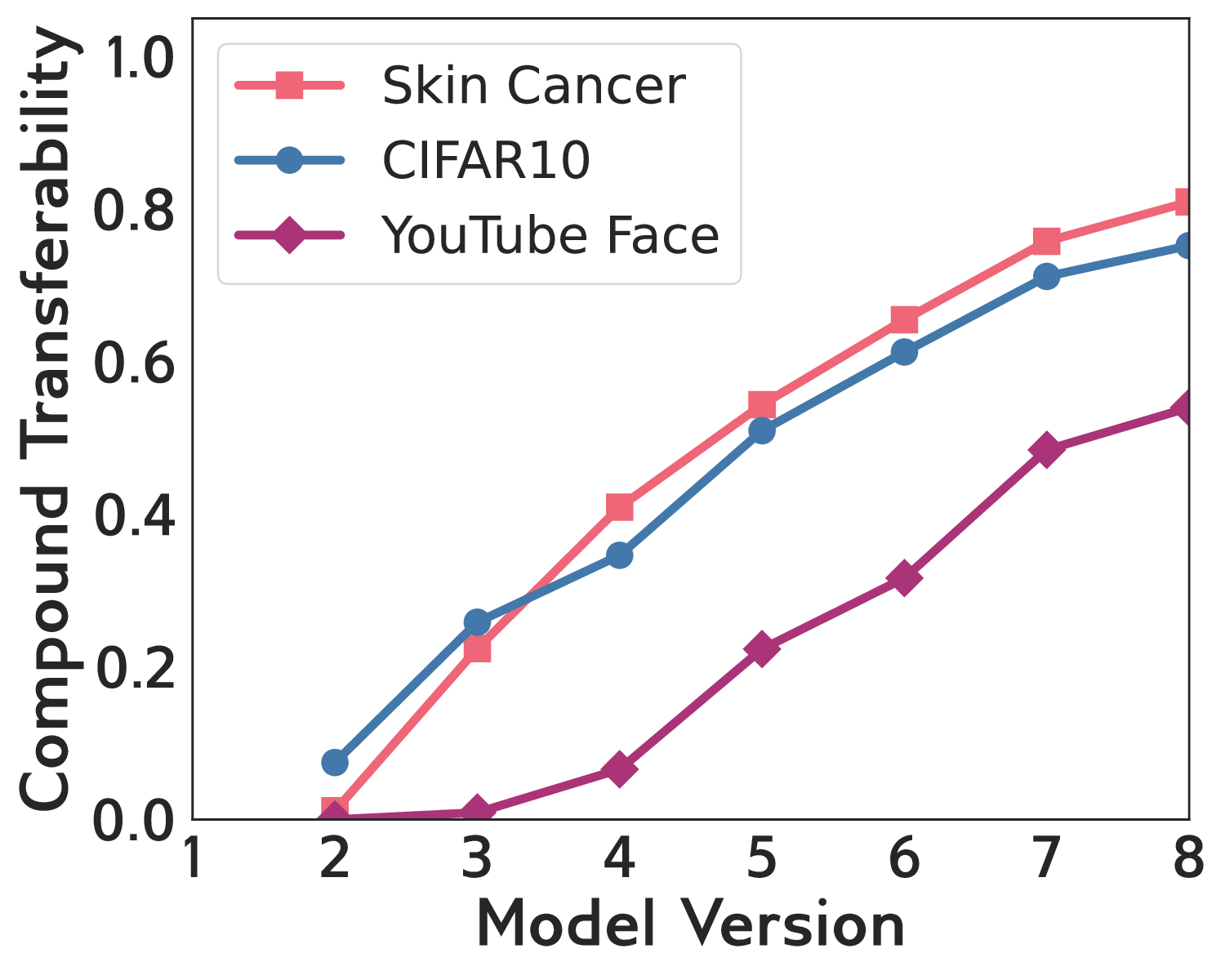}
	  \caption{\em Success rate of PGD-based compound transferability
            attacks against a sequence of $N$=8
            model versions.}
	  \label{fig:compound_hidden} 
    \end{minipage}
	\hspace{2pt}
	\centering
	\begin{minipage}[t]{0.31\textwidth}
	  \centering
	  \includegraphics[width=\textwidth]{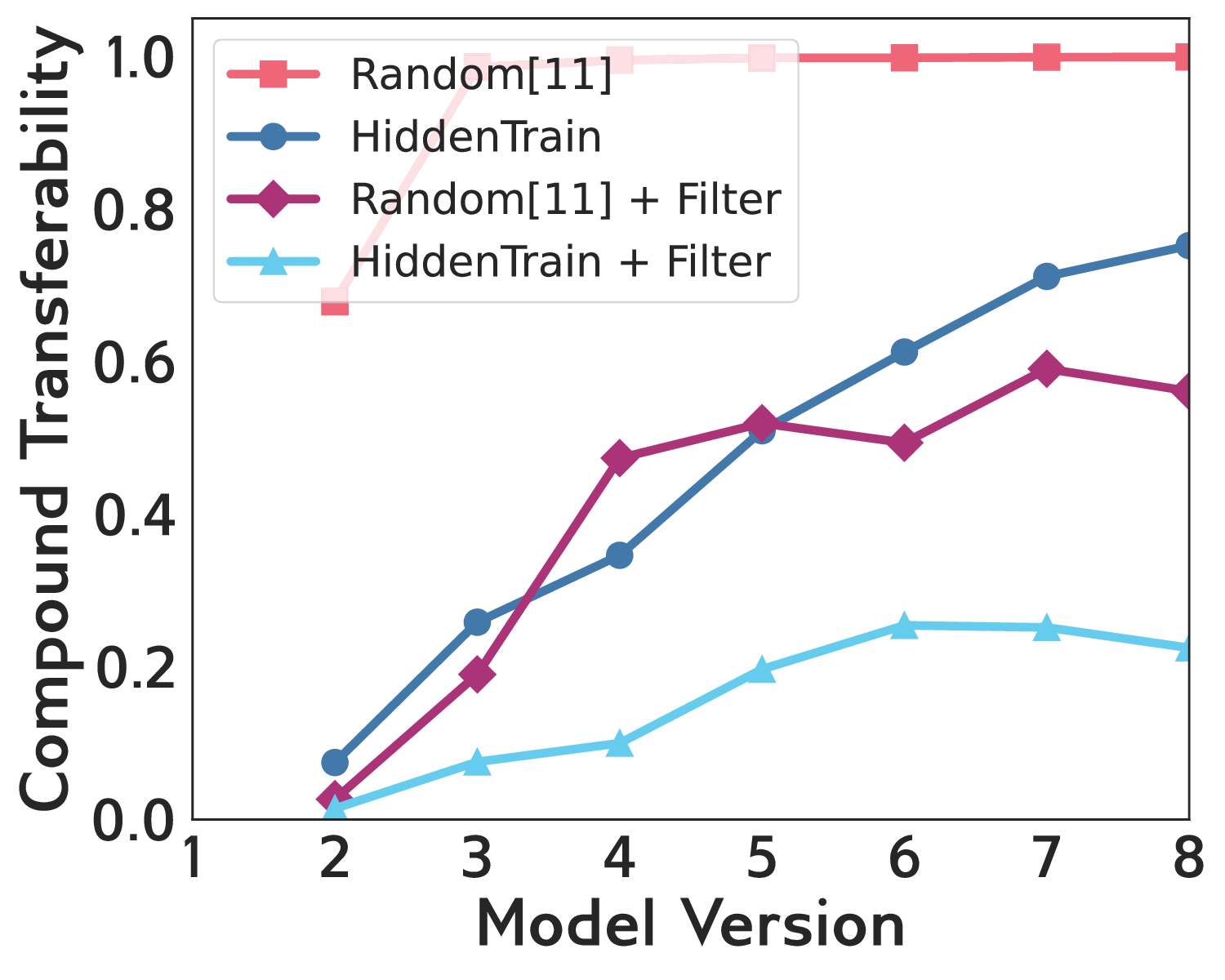}
        \caption{\em Success rate of PGD-based compound transferability attacks,
          ours (HiddenTrain) vs. [11] ($\cifar$).} 
        \label{fig:cifar_filter}
  \end{minipage}
  \hspace{2pt}
	\centering
  \begin{minipage}[t]{0.31\textwidth}
	  \centering
	  \includegraphics[width=\textwidth]{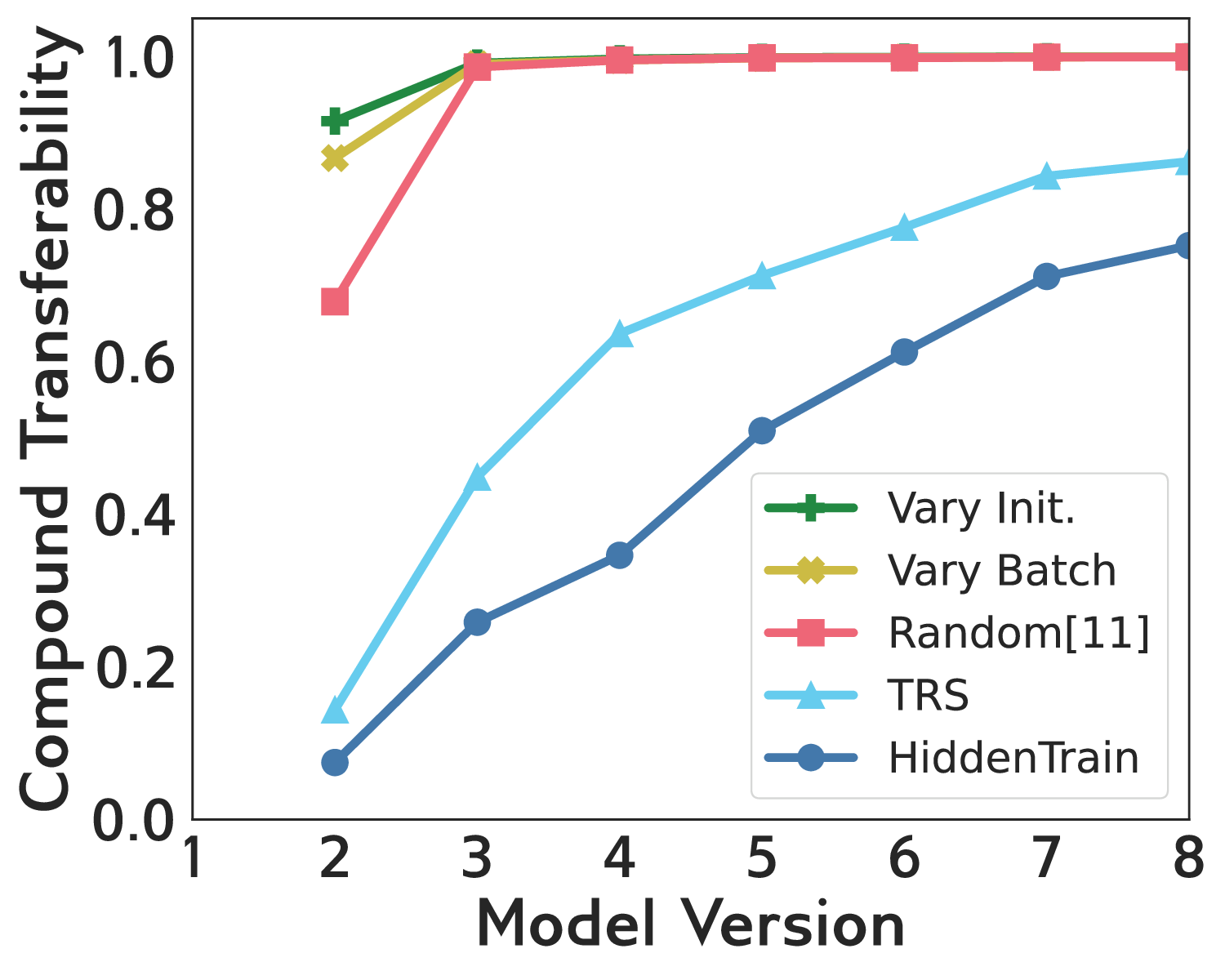}
	  \caption{\em Success rate of PGD-based compound transferability
            attacks under different versioning
            methods ($\cifar$).} 
	  \label{fig:alt}
  \end{minipage}
  \hfill
	\vspace{-0.2in}
\end{figure*}

\para{Robustness against Attacks.}  Next,
we evaluate the robustness of the model sequence created by hidden
training.  For each task and the pool of 50 model versions generated
by hidden training, we apply the greedy search method proposed in
\S\ref{sec:system} to build a sequence of $N=8$ models.

\vspace{3pt} \hspace{4pt} {\bf (1) Directional Transferability
  Attacks} --
Figure~\ref{fig:cifar_pair}-\ref{fig:ytface_pair} plot, for the
three classification tasks, the heatmap
views of the directional attack
transferability from a source model $j$ to a target model $i$ for 
the sequence of 8 models.   As previously mentioned, in our experiments, the attack
transferability equals the attack success rate. 

We see that the model versions maintain a
low directional transferability, even though our 
method aims at reducing the compound attack transferability.  The mean
directional transferability is low:  19\% for $\cifar$, 13\% for
$\skincancer$ and 2\% for $\ytface$.  It is worth
noting that, for all three tasks,  there is no obvious monotonicity in the directional
transferability as the model sequence grows. This means that the attacker is
unable to ``optimize'' the choice of model $j$ ($j<i$) to be used to attack model $i$.

\vspace{3pt} \hspace{4pt} {\bf (2) Compound Transferability Attacks} --  Next we
evaluate how the sequence of $8$ models resists the much stronger
compound transferability attacks, \revision{implemented using the ensemble
  attack proposed in~\cite{tramer2017ensemble}.

  We start from the
  case where the attacker launches an attack instance against the
  target model if it succeeds on at least one prior model. }  Figure~\ref{fig:compound_hidden} plots, for each of the three
classification tasks,  the attack success rate for
each model version in the sequence, which grows gracefully
with the model version count.  This trend aligns with our analytical
findings in \S\ref{sec:theory}. \revision{Moreover, we find that $90\%$ of
attack instances generated using the ensemble attack already succeed
on {\em all} previous model
	versions, demonstrating the powerfulness of the ensemble
        attack and the effectiveness of our method. 

 Next we consider ``cautious'' attackers who only deploy attack
 instances that succeed on all prior versions.  Table~\ref{tab:succeed_all}
	(Appendix~\ref{appendix:add_result}) shows that they only produce a minor increase in the compound attack
        transferability. Such increase can be effectively suppressed
        by combining hidden training with run-time input filtering
      proposed by~\cite{shan2022post} (discussed next).  }

\vspace{3pt} \hspace{4pt} {\bf (3) Hidden Training + 
  Run-time Detection} --  As a training-time defense, hidden training can be combined
with run-time attack detection systems to improve resilience against
compound transferability attacks.   To illustrate this, we implement, for 
each model version, the input filter proposed by~\cite{shan2022post}
to identify whether an input is an attack generated
from any of the previous model versions and
block any recognized as such. Under our scenario, these
detected attack inputs now carry zero
transferability. Figure~\ref{fig:cifar_filter} plots the attack success rate (in terms of
transferability) with and without the filter, for $\cifar$.  The
combined defense effectively suppresses the attack, \eg for model
version 8,  the attack success rate reduces drastically from 75.2\% to
22.5\%!  

To evaluate the ``contribution'' of hidden training to this combined
defense, we plot in the same figure the results of the model versioning
proposed by~\cite{shan2022post} with and without the input
filter.  We note that \cite{shan2022post} does not pick hidden
features but randomly selects a set of GAN-generated images as the
additional data to train model versions.   The
large gain over~\cite{shan2022post} demonstrates the
effectiveness of our hidden data selection process in building a more 
robust model sequence.  The increased diversity of models in the
sequence also contributes to enhancing the effectiveness of real-time attack detection.

\vspace{3pt} \hspace{4pt} {\bf (4) Comparison to Alternatives} -- We compare hidden training with
baseline techniques presented in \S\ref{subsec:goal}:
varying model initialization, varying training batch order, computing
model ensembles (TRS~\cite{yang2021trs}), and the random selection
method~\cite{shan2022post}.   Here we implement TRS~\cite{yang2021trs}
assuming that the model owner needs to train 8 models.  Due to the
extreme high
training cost of TRS (see \S\ref{subsec:time}), we were only able to
train the TRS models for $\cifar$.  Results in
Figure~\ref{fig:alt} show the success rate of compound
transferability attacks.  Here we can clearly observe the advantage of
hidden training in generating a robust model sequence. 
\label{subsec:comp_alt}

\subsection{Computation Overhead}
\label{subsec:time}
We examine computation overhead for hidden training and model
versioning.   Compared to training the original model,
generating a sequence of $N=8$ models requires (1) producing hidden
data required to train a pool of $50$ models,  (2) training  
$50$ models, and (3) running the greedy algorithm to select a sequence
of 8 models from the model pool.  We find that the computation overhead is dominated by model training.

  \begin{table}[h]
    \centering
    \begin{tabular}{ |c|c|c|c|c| } 
     \hline
           & Original &  \multicolumn{2}{c|}{Hidden Training}  &  TRS \\ \cline{3-4}
           & Model & (1 Model) & (50 Models) & (8 Models) \\
    \hline 
    $\cifar$ & $34.6$ s &  $34.8$ s & $29$ min & $> 100$ min \\ 
    $\skincancer$ & $193.5$ s  & $334$ s & $4.6$ hr &  $> 30$ hr\\ 
    $\ytface$ & $62.1$ s  & $112.5$ s & $1.5$ hr & $> 16$ hr\\ 
     \hline
    \end{tabular}
    \caption{\em Time spent to train a single epoch.}
    \vspace{-0.15in}
    \label{tab:time}
    \end{table}

Table~\ref{tab:time} lists the time required to train a single
epoch for the three classification tasks, using a Titan RTX GPU.  We also provide the training time for the original model and for TRS~\cite{yang2021trs}, assuming that the model owner sets up TRS to train 8 models as an ensemble.   These results show that our method consumes significantly less time compared to TRS, even when creating a pool of 50 models. For both $\skincancer$ and $\ytface$, we encountered convergence issues when attempting to produce 8 models using TRS, as each training epoch exceeded 16 hours in duration.

  \begin{table}[h]
	\centering
  \begin{tabular}{ |c|c|c| } 
   \hline
         &  Generate Hidden Data & Optimizing Sequence \\
  \hline 
  $\cifar$ & $70.8$ s & $< 6$ min \\ 
  $\skincancer$  & $824.8$ s & $< 15$ min\\ 
  $\ytface$  & $173.6$ s & $< 15$ min\\ 
   \hline
  \end{tabular}
  \caption{\em Computation overhead beyond model training. }
  \vspace{-0.1in}
  \label{tab:time_extra}
  \end{table}

Table~\ref{tab:time_extra} lists the additional overhead required
beyond model training, including time taken to generate and perturb
$2000$ images for a given feature point $\hiddenfeature$,  and average
time required to identify the next subsequent model.   For latter, the
main overhead is to generate adversarial examples using test data and estimate the compound attack transferability. 
This table shows that the time for generating hidden training
data per model is comparable to  2 epochs of hidden training,
indicating that the overall overhead is still dominated by
model training. The optimization time to produce a
sequence of $\totalversions$ models is less than 15 minutes$\times
(N-1)$, significantly less than the time required to train
$\totalversions=8$ TRS models.

\subsection{Ablation Study}
\label{subsec:ablation}
\begin{figure*}[t]
    \centering

    \begin{minipage}[t]{0.3\textwidth}
        \centering
        \includegraphics[width=\textwidth]{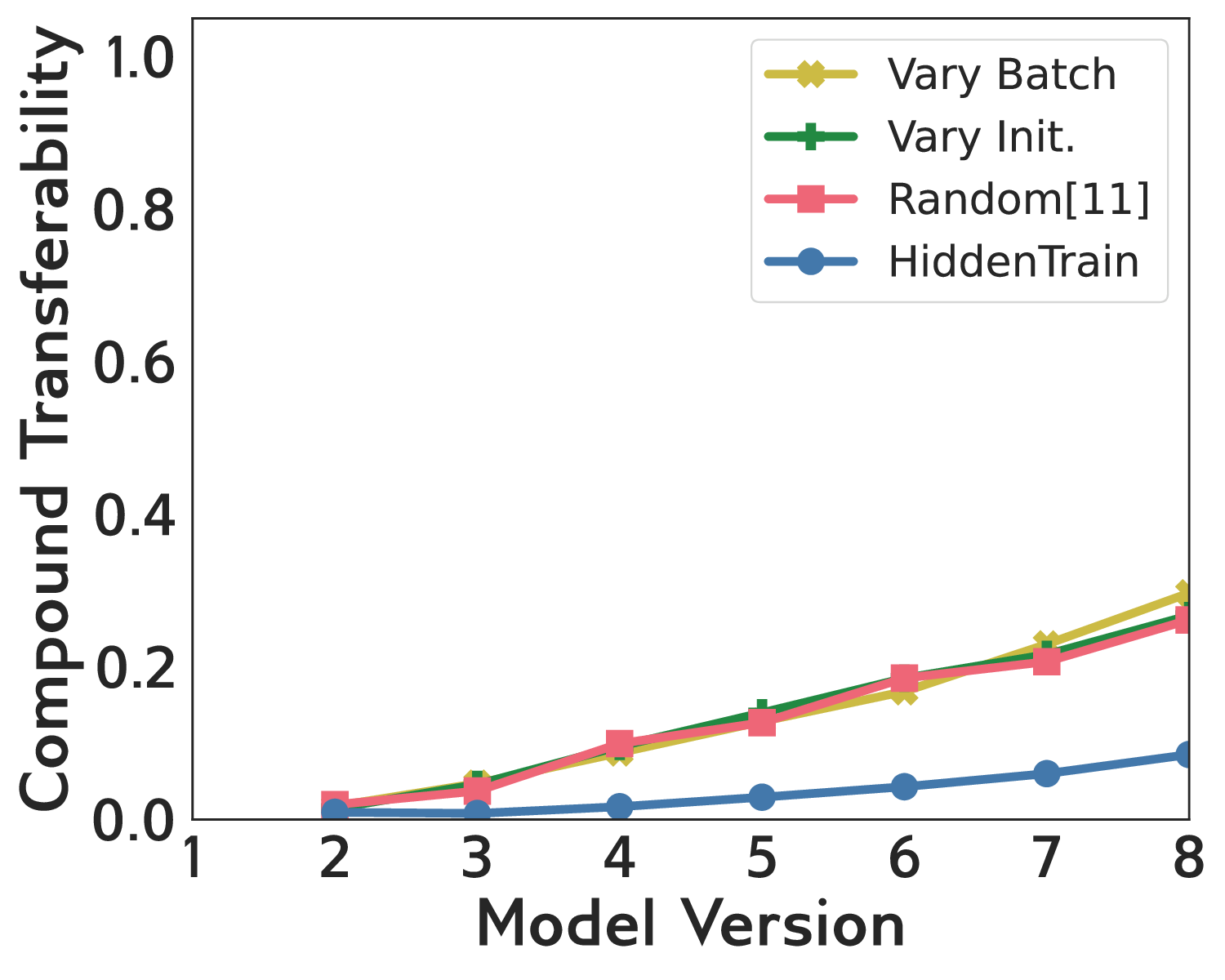}
        \caption{\em Success rate of CW-based compound transferability
        attacks under different versioning
        methods ($\cifar$)}
        \label{fig:cifar_cw}
    \end{minipage}
    \hfill   
    \hspace{7pt}
    \centering
    \begin{minipage}[t]{0.3\textwidth}
        \includegraphics[width=\textwidth]{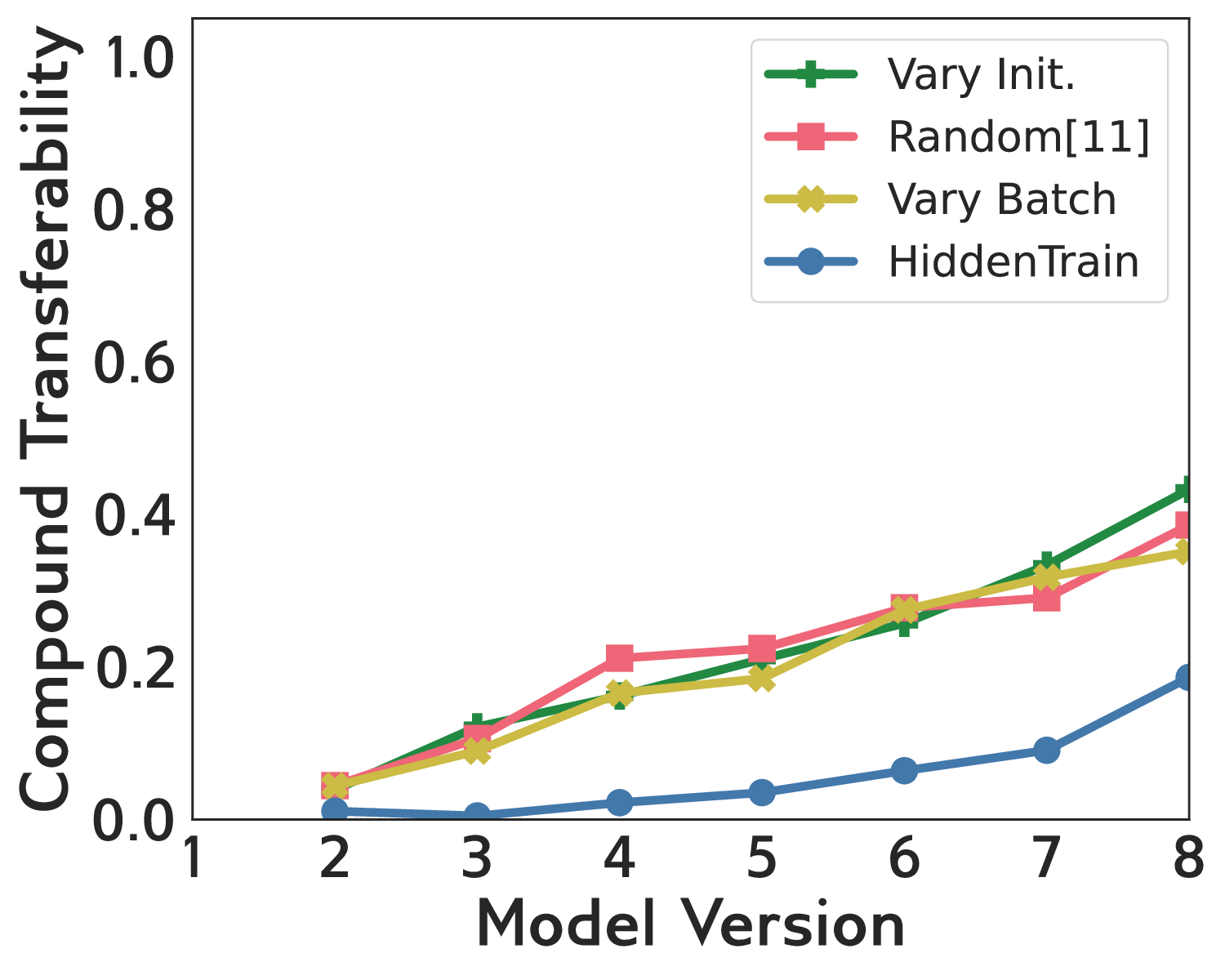}
        \caption{\em Success rate of EAD-based compound transferability
        attacks under different versioning
        methods ($\cifar$).} 
        \label{fig:cifar_eadl1}
    \end{minipage}
    \hfill   
  \hspace{7pt}
  \begin{minipage}[t]{0.3\textwidth}
    \includegraphics[width=\textwidth]{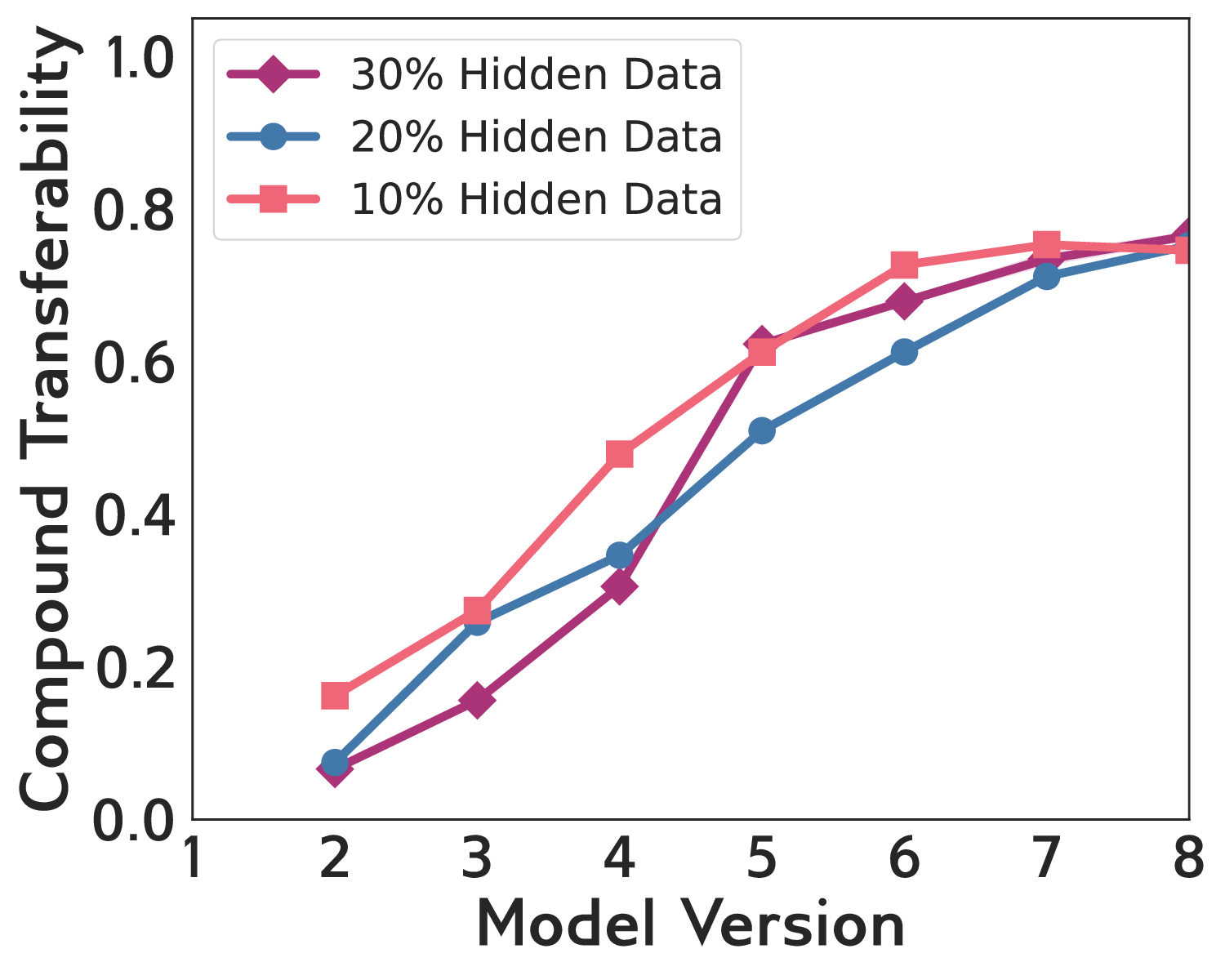}
    \caption{\em Success rate of PGD-based compound transferability
    attacks   under different the portion of hidden data
            ($\cifar$).} 
    \label{fig:cifar_changehidden}
\end{minipage}
\end{figure*}

\vspace{-6pt}
\para{Attack Configuration.} So far we report results assuming the
attacker launch PGD based attacks, which are known to carry strong
transferability across models. We also evaluate hidden training on two other
white-box adversarial attacks: CW~\cite{carlini2017towards} and
EAD~\cite{chen2018ead}. Figure~\ref{fig:cifar_cw}
and~\ref{fig:cifar_eadl1} (Appendix~\ref{appendix:add_result}) show the compound attack transferability of
multiple model versioning methods,  for both attacks.
Again, hidden training is the most effective at producing robust model sequences. 

\vspace{-1pt}
\para{Hidden Data Portion.} We also implement hidden training by
varying the portion of hidden data (relatively to the task training
data) from $10 \%$ to $30 \%$,  with $20 \%$ being the
default configuration in our experiments.
Results in Figure~\ref{fig:cifar_changehidden} (Appendix~\ref{appendix:add_result}) indicate that varying
the proportion of hidden data does not affect the effectiveness of
hidden training in maintaining a low compound attack
transferability. Furthermore,  all versions of the hidden-trained
models consistently achieve high normal classification accuracy
comparable to that of the original model (results omitted for
brevity). 

\vspace{-1pt}
\para{Impact of Model Architecture.} In addition to ResNet-18, we also
conduct hidden training using
VGG-16~\cite{simonyan2014very}. 
Figure~\ref{fig:cifar_vgg16} (Appendix~\ref{appendix:add_result}),
shows that hidden training yields the lowest compound attack
transferability,  demonstrating its applicability across multiple model
architectures. 

\vspace{-1pt}
\para{Protected Classes.}  We vary the number of protected
classes (see
Appendix~\ref{appendix:add_result}) and obtain consistent results:
model versioning via hidden training can protect multiple classes
against compound transferability attacks.

\section{Conclusion and Limitations}
\label{sec:discuss}
As classifiers are increasingly deployed in industrial settings, data breaches will inevitably
impact stored ML models, causing white-box model breaches. This
motivates us to address the pressing problem of robust model
versioning to maintain reliable ML services despite repeated model
leakages.  \revision{We introduce model versioning via hidden training, and demonstrate both
{\em theoretically} and {\em empirically} that, when properly
configured, they can produce model
versions robust against  multiple 
transferability-based attacks while achieving high task accuracy.}  Compared to alternative
methods such as random seeding and model ensembles, our method achieves higher
robustness (against compound transferability attacks), high scalability and
low cost for training model versions.

\vspace{-0pt}
\para{Limitations and Future Directions.} As the first work in this area, our work faces several limitations
that warrant additional research efforts.  First, our theoretical
analysis considers \revision{linear} SVM models. Further work is necessary to extend it
to DNNs.  Second, we propose a greedy method to
progressively construct DNN model versions from a pool of candidates,
demonstrating the feasibility and benefits of hidden
training. \revision{Yet the overhead for generating such pool of
  models can be heavy, especially for large models and for protecting
  many/all of the model classes simultaneously. Thus, we are unable to evaluate our design on large datasets like
  ImageNet.}   Additional
efforts are needed to produce stronger and more efficient optimization methods.  Finally, our experiments consider a loss-based,
compound attack to produce adversarial examples, \revision{defined by a prior
work~\cite{tramer2017ensemble}.}  Follow-up research should study the feasibility of
stronger attacks and refine the model versioning design to resist such
attacks.

\section*{Acknowledgements}
We thank the anonymous reviewers for their insightful feedback. We
thank Dr. Avrim Blum for his feedback on an earlier version of the
work, and Shawn Shan for his help on configuring the experiments.  
This work is supported in part by NSF grants CNS2241303 and
CNS1949650, the DARPA GARD program, and the C3.ai DTI program. Wenxin
Ding is supported by an Eckhardt Fellowship at the University of Chicago.  Any opinions, findings, and conclusions or recommendations expressed in this material are those of the authors and do not necessarily reflect the views of any funding agencies.

\bibliographystyle{IEEEtran}
\bibliography{IEEEabrv,hidden_dist}

\appendices
\onecolumn
In this appendix,  we list the detailed proofs for the three theorems presented 
\S\ref{sec:theory}.  We then present the hidden training algorithm in Appendix~\ref{appendix:hiddentrain}, and additional empirical results in Appendix~\ref{appendix:add_result}. 

\vspace{5pt}
\section{Proof of Theorem~\ref{thm:hidden_determine}}
\label{appendix:newSVM}
\begin{figure}[h]
    \centering
    \includegraphics[width=0.3\linewidth, angle=-90]{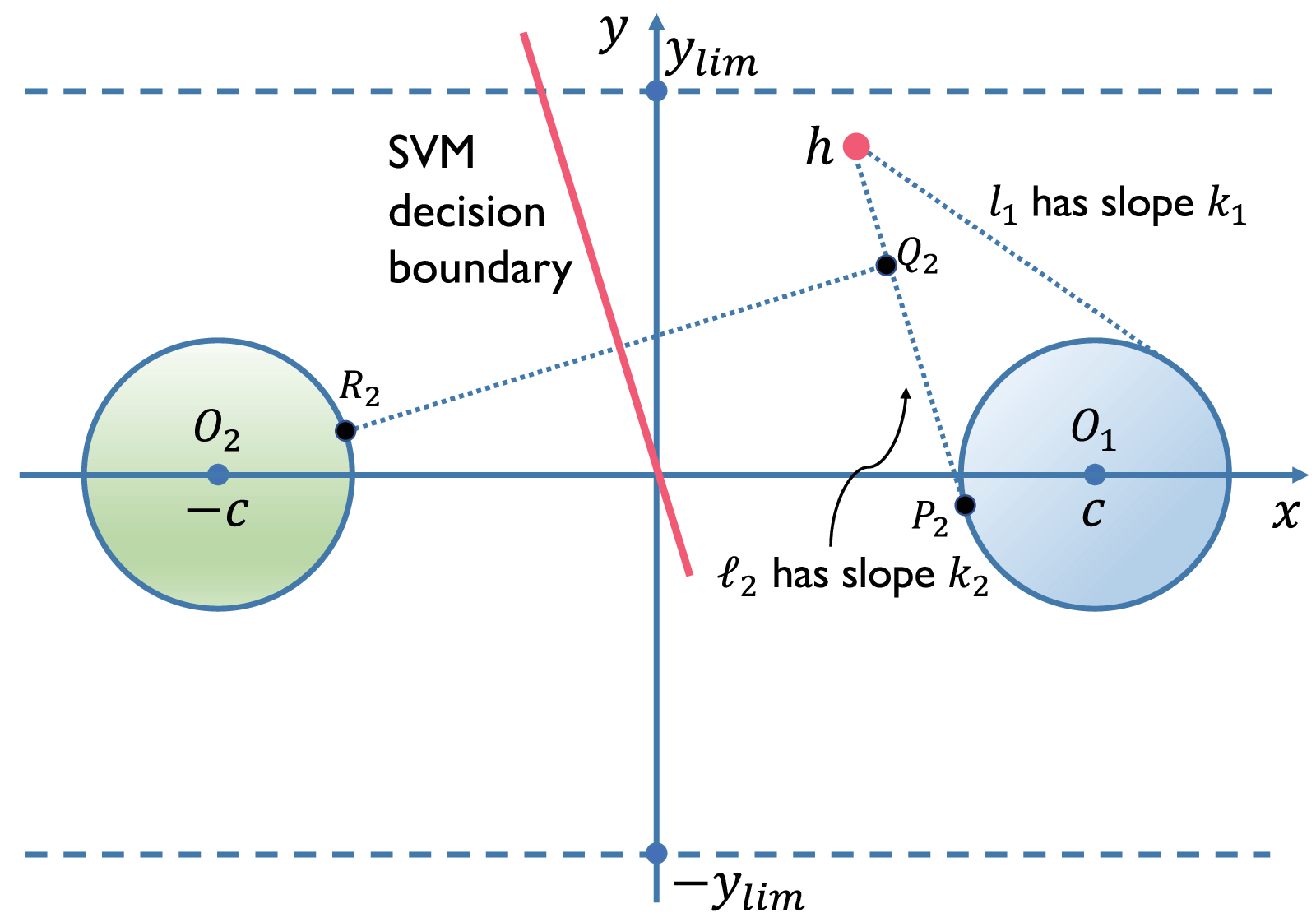}
    \caption{Illustration of an SVM decision boundary.
    }
    \label{fig:SVM_appendixA}
\end{figure}

In this proof, we explicitly compute the linear decision boundary of an SVM model in the presence of an added hidden data $h = (v,w)$. 
Without loss of generality, we assume $1-c\ < v < c-1$ and $|w| \le z$. We use Figure~\ref{fig:SVM_appendixA} to illustrate the notations and the process of finding the SVM decision boundary  described below. As stated earlier, in the SVM setting considered by our analysis, the input space and the feature space are identical.

We use $\mathcal{R}(p, q)$ to denote a unit circle centered at $(p,q)$, \ie $\mathcal{R}(p, q) = \{(x,y) \in \mathbb{R}^2 | (x-p)^2 + (y-q)^2 \le 1\}$.
As in the problem setup, we have two classes, $+$ and $-$. 
Training data for class $+$, denoted $\mathcal{D}_+$, is uniformly distributed in $\mathcal{R}(c, 0)$.  Training data $\mathcal{D}_-$ is uniformly distributed in $\mathcal{R}(-c, 0)$. Let $O_1$ denote a center $(c,0)$ and $O_2$ denote the other center $(-c,0)$, and we have $c>1$.

In a two-dimensional space, the optimal classifier is the one bisecting the shortest connection between the convex hulls of $\mathcal{D}_+$ and $\mathcal{D}_-$~\cite{scholkopf2002learning}. Therefore, we first find the convex hull with new training data $h$ in class $+$ and the other convex hull $\mathcal{R}(-c, 0)$ is not affected by $h$. We know that the convex hull of $\mathcal{D}_+ \cup \{h\}$ is enclosed by $\mathcal{R}(c, 0)$ and the line segments tangent to $\mathcal{R}(c, 0)$ passing through $h$.

To compute the lines tangent to $\mathcal{R}(c, 0)$, we assume the line is of the form $y=kx+b$. We then solve for $k$ and $b$ such that Equation~\ref{eq:tangent} only has one solution for $x$. Note that since $1-c\ < v < c-1$, no tangent line is of the form $x=b$.

\begin{equation}
    \label{eq:tangent}
    \begin{cases}
    & (x-c)^2 + y^2 = 1 \\
    & y=kx+b.
    \end{cases}
\end{equation}

This yields 
\begin{equation}
    \label{eq:quadratic}
    (x-c)^2 + (kx+b)^2 = 1  
\end{equation}
and the line being tangent indicates that the quadratic function in Equation~\ref{eq:quadratic} has only one solution for $x$.
Therefore, we solve for $k$ using
\begin{equation*}
    \begin{cases}
    & (2kb-2u)^2 -4(1+k^2) (u^2+b^2-1) = 0 \\
    &  w = kv+b.
    \end{cases} 
\end{equation*}

The tangent lines are denoted $\ell_1$ and $\ell_2$, have slopes $k_1 = \frac{-w(c-v) + \sqrt{(c-v)^2+w^2-1}}{(c-v)^2-1}$ and $k_2 = \frac{-w(c-v) - \sqrt{(c-v)^2+w^2-1}}{(c-v)^2-1}$, and have intercepts $w-k_1v$ and $w-k_2v$ respectively. Note that $k_1 > k_2$. Let $P_1$ and $P_2$ denote the tangent point on $\mathcal{R}(c, 0)$ by line $\ell_1$ and $\ell_2$ respectively.

If $w=0$, by symmetry of the convex hulls, the classifier is $x=\frac{-c+v+1}{2}$.

If $w > 0$, the shortest connection can be found by finding the shortest distance between $O_2$ and the line segment $P_2h$. Any point in the convex hull other than $P_2h$ has greater distance to $\mathcal{R}(-c, 0)$.

The orthogonal line to $\ell_2$ has slope $-\frac{1}{k_2}$. Let the line pass through $O_2$, we can then compute for its intercept. This line is of the form $y = -\frac{1}{k_2} (x-c)$.
Then we set the above computed line to intersect with $\ell_2$.
We denote the intersection point by $Q_2$.
$Q_2$ has coordinates $(\frac{k_2^2v - k_2w - c}{k_2^2 + 1}, \frac{-k_2 c - k_2 v + w}{k_2^2 + 1})$. 
If $\frac{k_2^2v - k_2w - c}{k_2^2 + 1} > v$ then the $Q_2$ is on the line segment $P_2h$. In this case $O_2Q_2$ is the shortest distance between $O_2$ and the convex hull. Otherwise, the shortest distance is $hO_2$.

If $Q_2$ is on the line segment $P_2h$, we find the point on $\mathcal{R}(-c, 0)$ that is intersected by $O_2Q_2$, denoted by $R_2$. We solve the equation
\begin{equation*}
    \begin{cases}
    & (x+c)^2 + y^2 = 1 \\
    & y = -\frac{1}{k_2} (x-c)
    \end{cases}
\end{equation*}
to get the coordinates of $R_2$ to be $(\frac{-k_2}{\sqrt{k_2^2+1}}-c, \frac{1}{\sqrt{k_2^2+1}})$. For the line bisecting $Q_2R_2$, it has the same slope as $\ell_2$. To find the intercept, we let the line pass through the mid-point of $Q_2R_2$.

Otherwise, we let $R_2$ be the point on $\mathcal{R}(-c, 0)$ that is intersected by $O_2h$. We solve the equation
\begin{equation*}
    \begin{cases}
    & (x+c)^2 + y^2 = 1 \\
    & y = \frac{w}{c+v} (x+c)
    \end{cases}
\end{equation*}
to get the coordinates of $R_2$ to be $(\frac{c+v}{\sqrt{(c+v)^2+w^2}}-c, \frac{w}{\sqrt{(c+v)^2+w^2}})$. We then bisect the line segment $R_2h$ to get the decision boundary.

Therefore,
\begin{itemize}
    \item If $\frac{k_2^2v - k_2w - c}{k_2^2 + 1} > v$, the shortest distance between the convex hulls is between $(\frac{k_2^2v - k_2w - c}{k_2^2 + 1}, \frac{-k_2 c - k_2 v + w}{k_2^2 + 1})$ and $(\frac{-k_2}{\sqrt{k_2^2+1}}-c, \frac{1}{\sqrt{k_2^2+1}})$. Bisecting the line, we can see that the decision boundary has slope $K(h) = k_2$ and intercept $B(h) = \frac{-k_2c+k_2v-w - \sqrt{k_2^2+1}}{2}$.
    \item Otherwise, the shortest distance between the convex hulls is between $(v, w)$ and $(\frac{c+v}{\sqrt{(c+v)^2+w^2}}-c, \frac{w}{\sqrt{(c+v)^2+w^2}})$. Bisecting the line, we can see that the decision boundary has slope $\frac{-c-v}{w}$ and intercept $\frac{-c^2+v^2+w^2 + \sqrt{(c+v)^2+w^2}}{2w}$.
\end{itemize}

When $w < 0$, by symmetry, we repeat the same analysis as in the case when $w > 0$. We omit some details in this case.

If $w < 0$, the shortest connection can be found by finding the shortest distance between $O_2$ and the line segment with slope $k_1$. The orthogonal line to $\ell_1$ passing through $O_2$ intersects $\ell_1$ at point $Q_1$ with coordinate $(\frac{k_1^2v - k_1w - c}{k_1^2 + 1}, \frac{-k_1 c - k_1 v + w}{k_1^2 + 1})$. If $\frac{k_1^2v - k_1w - c}{k_1^2 + 1} > v$ then the intersection is on the tangent line segment and $O_2Q_1$ is the shortest distance between $O_2$ and the convex hull. Otherwise, the shortest distance is $O_2P$.

\begin{itemize}
    \item If $\frac{k_1^2v - k_1w - c}{k_1^2 + 1} > v$, the shortest distance between the convex hulls is between $(\frac{k_1^2v - k_1w - c}{k_1^2 + 1}, \frac{-k_1 c - k_1 v + w}{k_1^2 + 1})$ and $(\frac{k_1}{\sqrt{k_1^2+1}}-c, -\frac{1}{\sqrt{k_1^2+1}})$. Bisecting the line, we can see that the decision boundary has slope $K(h)=k_1$ and intercept $B(h) = \frac{k_1c-k_1v+w - \sqrt{k_1^2+1}}{2}$.
    \item Otherwise, the shortest distance between the convex hulls is between $(v, w)$ and $(\frac{c+v}{\sqrt{(c+v)^2+w^2}}-c, \frac{w}{\sqrt{(c+v)^2+w^2}})$. Bisecting the line, we can see that the decision boundary has slope $K(h) = \frac{-c-v}{w}$ and intercept $B(h) = \frac{-c^2+v^2+w^2 + \sqrt{(c+v)^2+w^2}}{2w}$.
\end{itemize}

\vspace{5pt}
\section{Proof of Theorem~\ref{thm:no_transfer}}
\label{appendix:no_transfer}

First, we will show that if two SVM decision boundaries have opposite signs of slope and intersect at $(x_I, y_I)$ with $x_I \ge \delta$, then their attackable region has empty intersection. This yields zero directional attack transferability. We then explain how to find the hidden data $\hiddenfeature$ that satisfies the desired property.

Without loss of generality, let $\model_1$ have the decision boundary $y=k_1x + b_1$ with $k_1 < 0$ and $\model_2$ have the decision boundary $y=k_2x + b_2$ with $k_2 > 0$. Then for any $(x, y) \in \attackr_1$, we have $y \ge k_1 \delta + b_1$. Similarly, for any $(x, y) \in \attackr_1$, we have $y \le k_2 \delta + b_2$.

We can also compute the intersection of the two decision boundaries, where $x_I = \frac{b_2 - b_1}{k_1 - k_2}$. Since $x_I \ge \delta$, we have 
\begin{align*}
   & \frac{b_2 - b_1}{k_1 - k_2} \ge \delta \\
   \implies & b_2 - b_1 \le \delta \cdot (k_1 - k_2) \text{ since } k_1 < 0 < k_2 \\
   \implies & k_2 \delta + b_2 \le k_1 \delta + b_1.
\end{align*}
Therefore, $\attackr_1 \cap \attackr_2 = \emptyset$, which implies $\trans_{1 \to 2} = \trans_{2 \to 1} = 0$.

We now explain how to find $\hiddenfeature$ given a desired decision boundary $y=kx+b$. Without loss of generality, we assume $k > 0$.

\begin{figure}[h]
    \centering
    \includegraphics[width=0.3\linewidth, angle=-90]{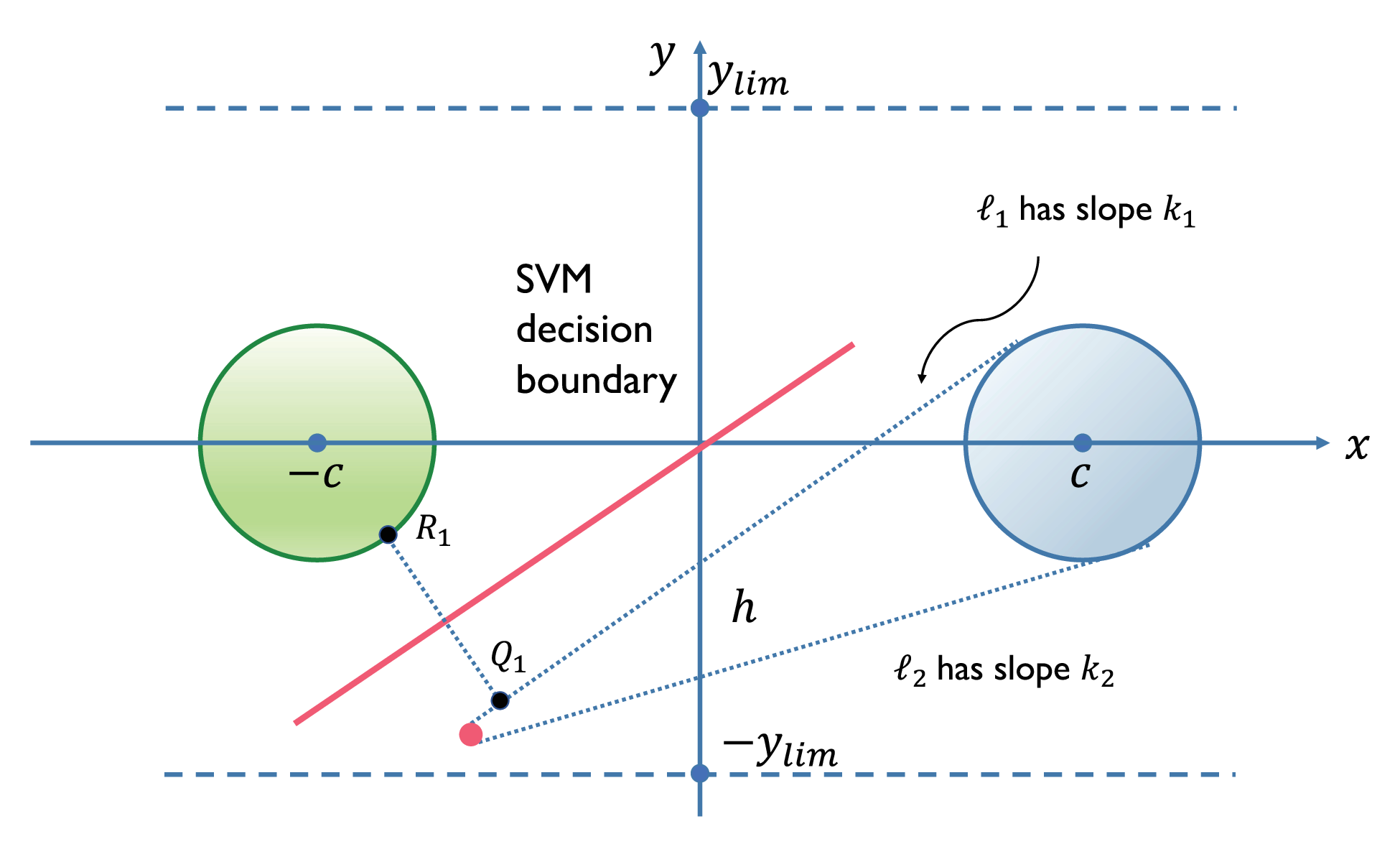}
    \caption{Illustration of a SVM decision boundary.
    }
    \label{fig:SVM_appendixB}
\end{figure}

By~\cite{scholkopf2002learning}, we know that the decision boundary has to bisect the convex hull of the two classes. Moreover, the convex hull of class $-$ is the unit ball at $(-c, 0)$. Therefore, we can find the function for the line that the decision boundary bisects, which is $R_1 Q_1$ in Figure~\ref{fig:SVM_appendixB}: $y = -\frac{1}{k} (x+c)$. Using this line, we can further find point $R_1$ by solving the equations
\begin{equation*}
    \begin{cases}
    & (x+c)^2 + y^2 = 1 \\
    & y = -\frac{1}{k} (x+c).
    \end{cases}
\end{equation*}
Solving the equations yields the coordinates of $R_1: (\frac{k}{k^2+1}-c, -\frac{1}{\sqrt{k^2+1}})$. The mid point of $R_1 Q_1$, which is also the intersection of the decision boundary with $R_1 Q_1$, has coordinates that satisfy
\begin{equation*}
    \begin{cases}
    & y = kx+b \\
    & y = -\frac{1}{k} (x+c).
    \end{cases}
\end{equation*}
Therefore, the mid point is $(\frac{-bk-c}{k^2+1}, \frac{b-ck}{k^2+1})$.

Using the coordinates of $R_1$ and mid point, we can find the coordinates of $Q_1$ because the mid point can also be found by $\frac{R_1 + Q_1}{2}$. Thus, $Q_1$ has coordinates $(\frac{2(-bk-c)}{k^2+1}-\frac{k}{\sqrt{k^2+1}}+c, \frac{2(b-ck)}{k^2+1} + \frac{1}{\sqrt{k^2+1}})$.

If $Q_1$ is a valid choice as $\hiddenfeature$, then we can choose $\hiddenfeature = Q_1$ and the resulting decision boundary will be the desired $y=kx+b$. The constraints are:
\begin{enumerate}
    \item $1-c < \frac{2(-bk-c)}{k^2+1}-\frac{k}{\sqrt{k^2+1}}+c < c-1$
    \item $|\frac{2(b-ck)}{k^2+1} + \frac{1}{\sqrt{k^2+1}}| \le y_{\lim}$
    \item $k_1 \cdot -\frac{1}{k} \ge -1$ \text{ where } $k_1$ \text{ is the larger slope of the line passing through } $Q_1$ { and tangent to the unit ball of class } $+$
\end{enumerate} 
The first two constraints enforce the $\hiddenfeature$ to be within the feasible region where we can add hidden data. The third constraint ensures that with $\hiddenfeature = Q_1$, the decision boundary is indeed the desired one. If $k_1 \cdot -\frac{1}{k} < -1$, then $R_1 Q_1$ is not the shortest distance between the convex hulls. Then the decision boundary would not bisect $R_1 Q_1$. Therefore, we can check the feasibility of the decision boundary using the constraints.

With the algorithm for checking feasibility, we can iteratively search through the space to find a feasible linear separator.

\vspace{5pt}
\section{Proof of Theorem~\ref{thm:sequence}}
\label{appendix:compound}
In Appendix~\ref{appendix:no_transfer}, we discuss how we validate feasibility of an SVM decision boundary. In this section, we explain how we construct a sequence of SVM models such that we can upper bound the maximum compound transferability of the sequence.

\para{When $\totalversions = 2$:}
We build $\model_1$ by selecting some $k>0$ such that $y=k(x-\delta)$ is a feasible decision boundary,  using the verification method discussed in Appendix~\ref{appendix:no_transfer}. Since  $y=k(x-\delta)$ is a feasible decision boundary, then by symmetry, $y=-k(x-\delta)$ is also feasible. Therefore, we can build $\model_1$ with decision boundary $y=k(x-\delta)$ and $\model_2$ with decision boundary $y=-k(x-\delta)$. By Appendix~\ref{appendix:no_transfer}, we have $\attackr_1 \cap \attackr_2 = \empty$ so $\trans(\model_1 \rightarrow \model_2) = \trans(\model_2 \rightarrow \model_1) = 0$.

\para{When $\totalversions > 2$:} In this general case, we also start by finding a feasible decision boundary $y=k(x-\delta)$. 
Ideally, a larger $k$ results in a smaller attackable region. However, since we need to choose a sequence of models with large separations,  the chosen $k$ cannot be too large.  After selecting a feasible $k$ value, we seek to find the largest $b>0$ such that the decision boundary $y=kx-b$ remains feasible.  We denote this maximum value of $b$ as $b_{\max}$.   Here we observe that as $k$ increases, the value of $b_{max}$ satisfying the above requirement decreases.

Next we set $n = \lceil \frac{\totalversions}{2}  \rceil - 1$ and $b = \frac{b_{\max}}{n}$.  We set $\model_1$ with decision boundary $y = k(x-\delta)$ and $\model_2$ with $y = -k(x-\delta)$. For $\model_i$ ($i > 2$), if $i$ is odd, we set its decision boundary to $y = k(x-\delta) - b \cdot \frac{i-1}{2}$;  if $i$ is even, we set its decision boundary to $y = -k(x-\delta) + b \cdot \frac{i-2}{2}$.

As $i$ increases, $\attackr_i$ decreases under our construction. Moreover, $\cup_{j=1}^{i-1} \attackr_j = \attackr_1 \cup \attackr_2$ for all $i>2$. Therefore, the compound transferability of the sequence is bounded by $\alpha_\totalversions$ where $\alpha_\totalversions = \trans(\{\model_j\}_{j=1}^2 \rightarrow \model_3)$. This ends our proof.

\para{Specific Realization Shown in Table 1.}  As an illustrative example, we assume $c=100$, $y_{lim}=30$ and $\delta = 0.1$. We choose $k=7$ and select $b_{\max}=12$. Note that actual $b_{\max}$ is slightly greater than $12$,  but we choose $12$ to simplify the computation.

For an SVM with decision boundary $y=kx-b$ satisfying $k>0$, $b>0$ and $\frac{b-y_{lim}}{k} < -\delta$, we have $\areafunc(\attackr) = \frac{(y_{lim}-k \cdot \delta -b)^2}{2k} + \delta \cdot (y_{lim}-b + \frac{k\delta}{2})$. This is computed by summing up the area of a triangle and a trapezoid, as shown in the shadowed area in Figure~\ref{fig:SVM}.  Given the above configuration, we have $\areafunc(\attackr_1) = \areafunc(\attackr_2) = 61.39$,  with $\attackr_1 \cap \attackr_2 = \emptyset$.

Next, we compute $\alpha_\totalversions$ for each given $N$ value.  When $3 \le \totalversions \le 4$, we have $b = 12$ and $\alpha_\totalversions = \frac{\areafunc(\attackr_3)}{2\areafunc(\attackr_1)} = 0.17$. When $5 \le \totalversions \le 6$, we have $b = 6$ and $\alpha_\totalversions = 0.32$. When $7 \le \totalversions \le 8$, we have $b = 4$ and $\alpha_\totalversions = 0.37$. And finally when $9 \le \totalversions \le 10$, we have $b = 3$ and $\alpha_\totalversions = 0.4$.

\vspace{5pt}
\section{Hidden Training Algorithm}
\label{appendix:hiddentrain}

In the following, we provide a detailed discussion on our hidden training algorithm, in the context of producing a sequence of models.   This is done in four steps.

\para{Step 1: Selecting a candidate set of feature points (Algorithm~\ref{alg:hidden_feature}).} 
Our analytical study in Appendix~\ref{appendix:no_transfer} and Appendix~\ref{appendix:compound} shows that effective hidden feature points often lie at the edge of the classes. Therefore, we first examine the original feature space and choose the feature vectors that are located at a distance from the center of each original class. Here we face two constraints: (1) features in the original feature space that are too far from the original classes may not be realizable in the input space, \ie it is difficult to generate the corresponding input data, and (2) the chosen feature points should not overlap with feature vectors of the task classes.  These two considerations lead us to apply a feature-multiplier $m$ to ``drag'' existing feature vectors away from the original classes, and bound $m$ by $1.1 < m < 1.5$.  As such, Algorithm~\ref{alg:hidden_feature} outputs a list of chosen feature points.

\para{Step 2: Generating hidden data in the input space (Algorithm~\ref{alg:hidden_train}).} 
Next, we generate a set of hidden data (in the input space) from each chosen feature point $h$ and use them to train a model version parameterized by $h$. For this, we apply the algorithm from~\cite{shafahi2018poison}, which perturbs a given input $x$ to move its feature vector to the target feature vector in the original feature space.  Here we use a pre-trained GAN model~\cite{karras2017progressive} to generate a set of $2000$ images as the initial inputs,  and perturb each input to reach the chosen $h$.
For each $\hiddenfeature$, we apply up to $100$ perturbation iterations (without any perturbation budget) to produce the corresponding hidden data.  We then split the hidden data into training and testing sets,  where the hidden data used for training is 20\% of the original training data.  The testing hidden data is used to validate the model performance.

\begin{figure*}[h]
    \centering
    \begin{minipage}{0.3\textwidth}
		\centering
		\includegraphics[width=\textwidth]{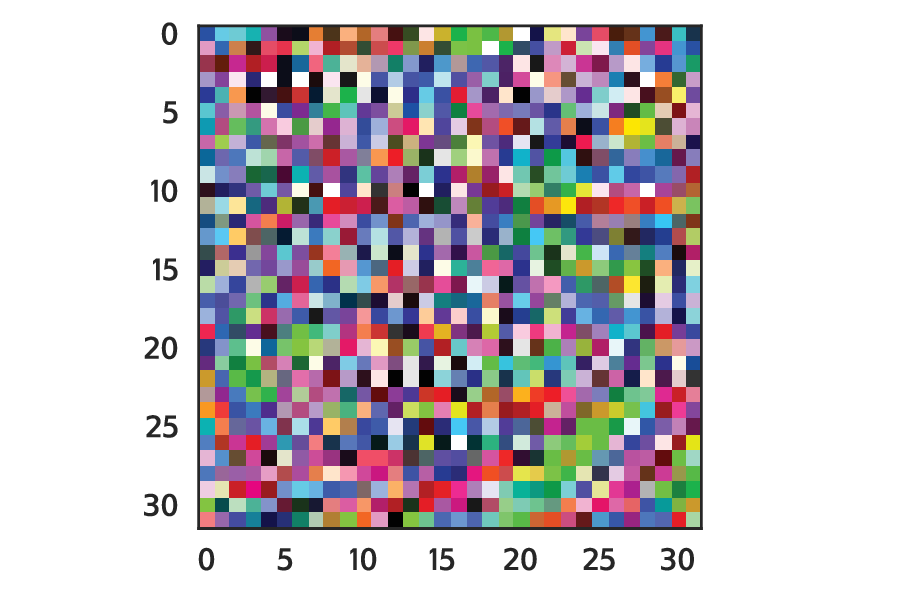}
		\caption{\em Sample hidden data used for $\cifar$.}
		\label{fig:sample_cifar}
	\end{minipage}
    \hspace{7pt}
    \centering
    \begin{minipage}{0.3\textwidth}
        \centering
        \includegraphics[width=\textwidth]{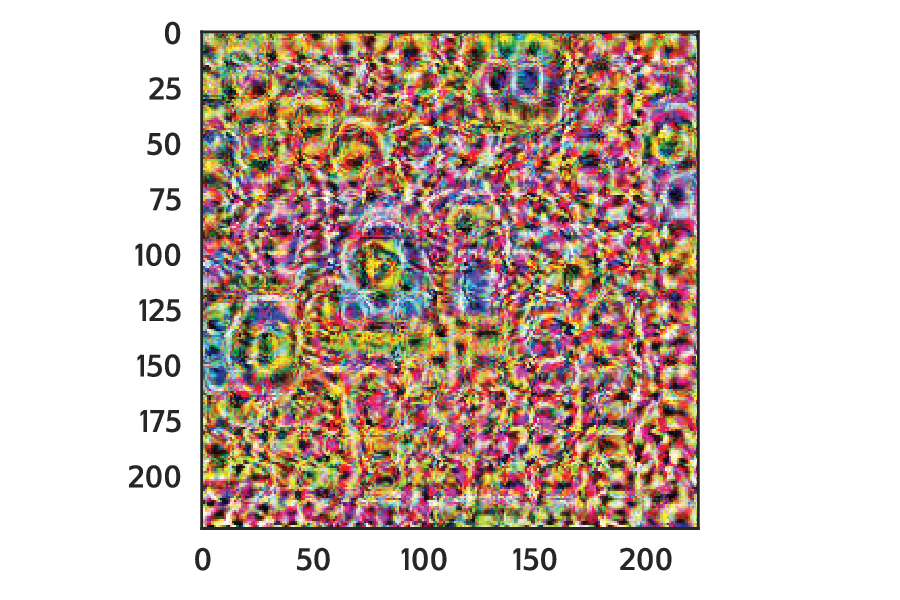}
        \caption{\em Sample hidden data used for $\skincancer$.}
        \label{fig:sample_skincancer}
    \end{minipage}
    \hspace{7pt}
    \centering
    \begin{minipage}{0.3\textwidth}
		\centering
		\includegraphics[width=\textwidth]{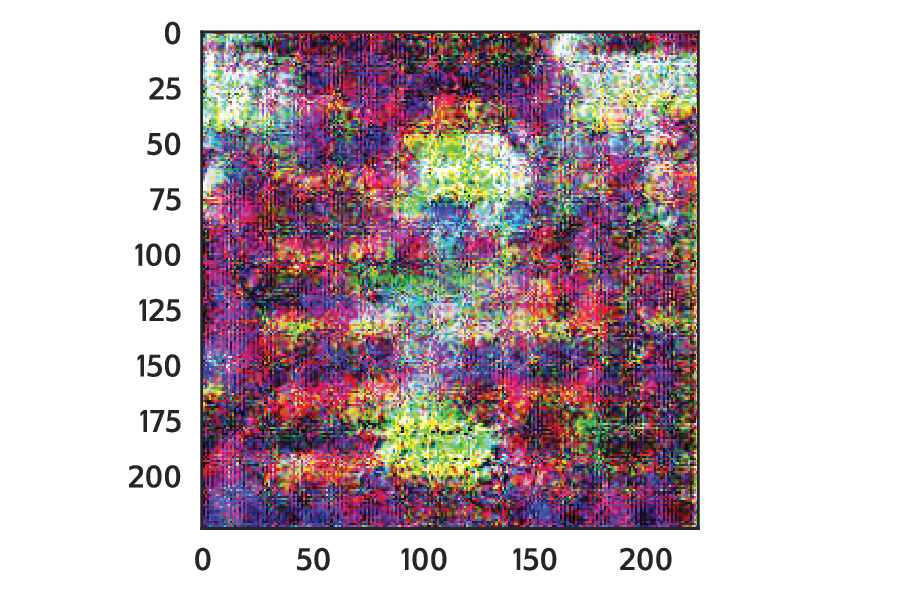}
		\caption{\em Sample hidden data used for $\ytface$.}
		\label{fig:sample_ytface}
	\end{minipage}	
    \hfill
\end{figure*}

Later in Figure~\ref{fig:sample_cifar},~\ref{fig:sample_skincancer} and~\ref{fig:sample_ytface}, we provide sample images of the hidden data generated for the three classification tasks used in our experiments. We would like to emphasize here that the key property of the hidden data is not what it looks like in the input space, but rather how close its representations are to the chosen feature point in feature space. For other application domains, the visual representation is a moot point.

\para{Step 3: Training candidate model versions.}  Now given the hidden data produced in Step 2, we can run the usual model training by adding these data to the training set.  This step creates a set of candidate model versions, referred to as $\mathcal{S}_\model$. 

\para{Step 4: Configuring the model sequence (Algorithm~\ref{alg:optimization}).}  Given the model pool $\mathcal{S}_\model$, we apply a greedy search method to choose the sequence of models $\model_1, \model_2, ...  \model_{i}, ... $.  The goal is to find $\model \in \mathcal{S}_\model$ that has the lowest compound attack transferability from $\model_1, ..., \model_{i-1}$. Ideally, we would like to search the whole feature space to find the feature point $\hiddenfeature_i$ that minimizes compound transferability. However, such global optimization is very difficult because the impact of $\hiddenfeature_i$ on the trained model version cannot be explicitly formulated. 
Instead, we apply a practical, greedy search method to gradually choose from the pool the next version $i$. The selection is driven by computing the compound transferability attacks and launching them on the candidate models to estimate the compound transferability.  The candidate model with the lowest transferability is then selected.  Therefore, the larger the pool size, the better the optimization. For our implementation, the pool size is $50$ to achieve a low computation overhead.

\begin{algorithm}
    \caption{Finding Feature Points}\label{alg:hidden_feature}
    \begin{algorithmic}
    \State \textbf{Input}: Original model $\modelori$, test data $\testori$, class labels $\mathcal{L}$, distance metric $d$, distance threshold $\epsilon_d$, multiplier $m$
    \State \textbf{Output}: a list of $\totalversions$ feature points
    \State Feature points $\mathcal{H} \gets \{\}$
    \State $\featurespaceori \gets $ feature extractor from $\modelori$
    \For {$\ell \in \mathcal{L}$}
        \State $X_\ell = \{(x, c) \in \testori \; | \; c=\ell\}$
        \State $F_\ell = \featurespaceori(X_c)$
        \State $h_\ell = mean(F_\ell)$
        \For {$h \in F_\ell$}
            \If {${d(m*h, h_\ell) > \epsilon_d}$}
                \State Add $m*h$ to $\mathcal{H}$
            \EndIf
        \EndFor
    \EndFor
    \State Output $\mathcal{H}$
\end{algorithmic}
\end{algorithm}

\begin{algorithm}[H]
    \caption{Hidden Training }\label{alg:hidden_train} 
    \begin{algorithmic}
    \State \textbf{Input}: Original model $\modelori$, training data $\trainori$,   set of target classes $\mathcal{L}_T$, list of GAN images $\mathcal{S}_\mathcal{G}$, list of feature points $\mathcal{S}_\mathcal{H}$, perturbation algorithm from~\cite{shafahi2018poison} $Perturb(.)$
    \State \textbf{Output}: a list of hidden trained model versions $\mathcal{S}_\model$
    \State Model list $\mathcal{S}_\model \gets \{\}$ 
    \State Number of protected classes $n_t \gets len(\mathcal{L}_T)$
    \For {$i$ in $(0, len(\mathcal{S}_\mathcal{H}), n_t)$}
        \State $\trainhidden_i \gets \{\}$
        \For {$j, l_t$ in $enumerate(\mathcal{L}_T)$}
            \State $g \gets \mathcal{S}_\mathcal{G}[i*n_t + j]$
            \State $h \gets \mathcal{S}_\mathcal{H}[i*n_t + j]$
            \State $X_h \gets Perturb(\modelori, h, g)$
            \State $\trainhidden_i = \trainhidden_i \cup \{(x, l_t) \; | \; x \in X_h\}$
        \EndFor
        \State Train $\model_i$ with $D_{\text{train}} \cup \trainhidden_i$
        \State Add $\model_i$ to $\mathcal{S}_\model$
    \EndFor 
    \State Output $\mathcal{S}_\model$
\end{algorithmic}
\end{algorithm}

\begin{algorithm}[H]
    \caption{Forming Model Sequence by Greedy Search}\label{alg:optimization}
    \begin{algorithmic}
    \State \textbf{Input}: List of hidden trained models $\mathcal{S}_\model$, list of breached models $\mathcal{B}_\model$, adversarial example generating algorithm $Adv$, transferability computation algorithm $Trans(.)$, training data $\testori$, set of target classes $L_T$
    \State \textbf{Output}: a model to be deployed
    \For{ $l_t$ in $enumerate(L_T)$}
    \State $X_{adv} = X_{adv} \cup Adv(\mathcal{B}_\model, \testori, l_t)$
     \EndFor
    \State $\model = \argmin_{\model_i \in \mathcal{S}_\model, \model_i \notin \mathcal{B}_\model} Trans(\model_i, X_{adv})$
    \State Output $\model$
\end{algorithmic}
\end{algorithm}

\section{Training and Attack Configurations}
\label{appendix:config}

\revision{In this section, we add additional information on the training and attack configurations. The configurations are listed in Table~\ref{tab:config}. We show the training configuration of the models in the original results and the attack configurations of PGD attacks.}

\begin{table}[h]
    \centering
    \begin{tabular}{ |l|l|l|l|l|l|l|l|l| } 
        \hline
         & \multicolumn{5}{c|}{Training} & \multicolumn{3}{c|}{Attack} \\
         \cline{2-9} 
        & Architecture & Epochs & Batch Size & Optimizer & \makecell{Learning \\Rate} & \makecell{Perturbation\\ Budget} & Iterations & $\alpha$ \\
        \hline
        $\cifar$ & ResNet-18 & 20 & 512 & SGD & 0.5 & 0.03 & 30 & 0.01 \\ 
          $\skincancer$ & Densenet-121 & 10 & 64 & Adam & $1e-3$ & 0.05 & 30 & 0.01\\ 
          $\ytface$ & ResNet-50 & 20 & 32 & Adam & $1e-4$ & 0.25 & 100 & 0.05 \\         
        \hline
    \end{tabular}
    \caption{\em Training and attack configurations.}
    \vspace{-0.1in}
    \label{tab:config}
    \end{table}

\section{Additional Results on Robustness}
\label{appendix:add_result}

In this section, we provide additional results for ablation study on the impact of model architecture and the number of protected classes.

\begin{figure*}[t]
    \centering
    \begin{minipage}{0.3\textwidth}
		\includegraphics[width=\textwidth]{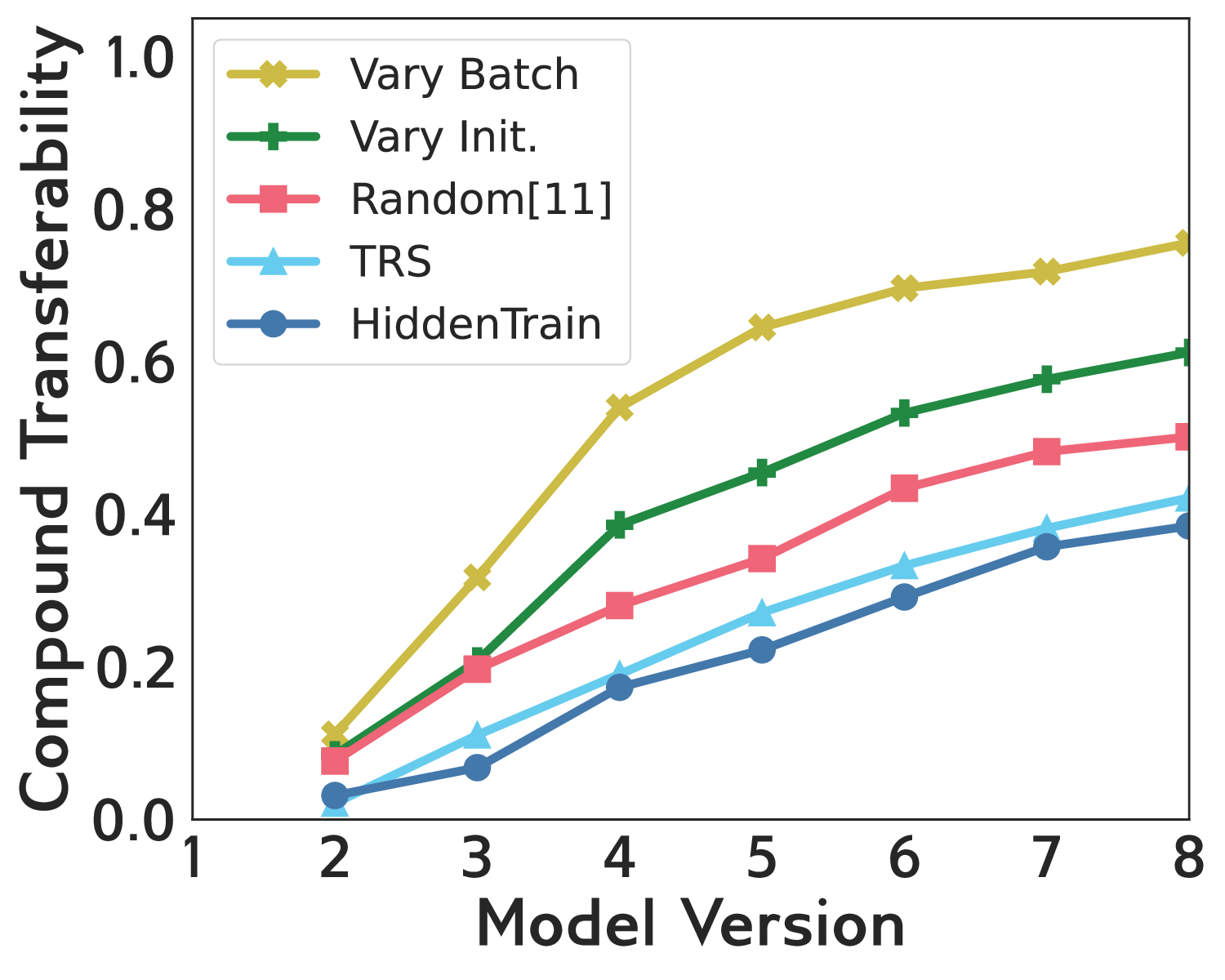}
		\caption{\em Success rate of PGD-based compound transferability
        attacks under different versioning
        methods ($\cifar$, VGG16 model architecture).} 
		\label{fig:cifar_vgg16}
	\end{minipage}
    \hspace{7pt}
    \centering
    \begin{minipage}{0.3\textwidth}
		\resizebox{1\textwidth}{!}
        {
            \begin{tabular}{ l|l|l|l|l|l|l|l } 
                \hline
                \makecell{Attack \\ Instance \\ Selection}& $2$ & $3$ & $4$ & $5$ & $6$ & $7$ & $8$ \\
                \hline
                \multicolumn{8}{c}{Without Filter} \\ 
                \hline 
                \makecell{Succeed on \\ at least 1 \\ prior model} & $0.07$ & $0.26$ & $0.35$ & $0.51$ & $0.61$ & $0.71$ & $0.75$\\ 
                \hline
                \makecell{Succeed on all \\prior models} & $0.07$ & $0.27$ & $0.36$ & $0.54$ & $0.66$ & $0.77$ &$0.84$\\ 
                \hline
                \multicolumn{8}{c}{With Filter} \\ \hline
                \makecell{Succeed on \\ at least 1 \\ prior model \\ {\em with} Filter} & $0.01$ & $0.07$ & $0.10$ &$0.20$ & $0.25$ & $0.25$ & $0.22$\\ 
                \hline
              \makecell{Succeed on all \\prior models \\ {\em with} Filter} & $0.02$ & $0.06$ & $0.10$ & $0.21$ & $0.25$ & $0.27$ & $0.28$\\
                \hline
            \end{tabular}
        }
		\captionof{table}{\revision{\em Success rate of PGD-based compound transferability
        attacks from a ``cautious'' attacker, with and without run-time attack detection and filtering ($\cifar$).}} 
		\label{tab:succeed_all}
	\end{minipage}
    \hspace{7pt}
    \centering
    \begin{minipage}{0.3\textwidth}
        \centering
        \resizebox{1\textwidth}{!}{
            \begin{tabular}{ l|l|l } 
                \hline
                & \makecell{Protect \\ 1 class} & \makecell{Protect \\ 3 classes}  \\
                \hline 
                \makecell{Random selection~\cite{shan2022post} \\{\em without} Filter} & $99.93 \%$ & $99.38 \%$  \\ \hline
                \makecell{Hidden Training \\ {\em without} Filter} & $75.22 \%$ & $85.75 \%$ \\ \hline
                \makecell{Random selection~\cite{shan2022post} \\ {\em with} Filter} & $56.14 \%$ & $49.14 \%$ \\ \hline
              \makecell{Hidden Training \\ {\em with} Filter} & ${\bf 22.5} \%$ & ${\bf 26.95} \%$\\
                \hline
            \end{tabular}
        }
        \captionof{table}{\em Success rate of PGD-based compound transferability
        attacks against the $8^{th}$ model version in the sequence, under different number of protected classes ( $\cifar$).}
        \label{tab:protect_multi}
    \end{minipage}
    \hfill   
\end{figure*}

\para{Impact of Model Architecture} We train VGG-16 models for the $\cifar$ dataset and study the compound attack transferability for a sequence of 8 models. Figure~\ref{fig:cifar_vgg16} plots the success rate of compound transferability attacks, for both hidden training and four alternative methods. Again hidden training outperforms its alternatives. This result is consistent with prior results using ResNet18 models,  demonstrating the applicability of hidden training across multiple model architectures.

\para{Protected Classes.} We also vary the number of protected classes when applying hidden training to create the model sequence. In Table~\ref{tab:protect_multi}, we present the compound attack transferability experienced by the 8th model in the sequence.  Maintaining robustness at this model version is the hardest among the eight model versions, as the attacker now possesses white-box access to all preceding seven models. Additionally, we also include the results when combining hidden training with run-time attack detection and filtering, along with the results of using random hidden feature selection~\cite{shan2022post}.

The results in Table~\ref{tab:protect_multi} show that protecting either 1 or 3 classes yields similar robustness performance. Furthermore, combining hidden training with run-time attack detection is highly effective in sustaining robustness even when the requirement is to protect a larger number of classes.  As expected, hidden training largely outperforms random selection~\cite{shan2022post}, with or without run-time filtering.  Together, these findings demonstrate the ability of hidden training to safeguard multiple classes simultaneously.

\revision{\para{Attack Aggressiveness.} We also consider an attacker that choose to submit an attack instance to model $\model_i$ only if it succeeds on all prior models. We consider such attacks as cautious attacks. In our original experiments, we consider attacks that succeed on at least one previous models. When we examine the attack instances used by our original experiments, we find that  more than $90\%$ of the instances already succeed on all previous models.

Table~\ref{tab:succeed_all} shows the result of comparing compound transferability with attacks selected using the above described two criteria. We see that the compound transferability of cautious attacks is slightly higher, especially at later model versions. However,  the change is minor, especially when compared to~\cite{shan2022post}, whose transferability is more than $90\%$ beyond version 2.}

\end{document}